\documentclass[10pt,twocolumn,letterpaper]{article}

\usepackage{epsfig}
\usepackage{graphicx}
\usepackage{amsmath}
\usepackage{amssymb}
\usepackage{adjustbox}
\usepackage{hyperref}

\begin{document}

\title{Towards Streaming Egocentric Action Anticipation}

\author{Antonino Furnari\\
University of Catania\\
{\tt\small furnari@dmi.unict.it}
\and
Giovanni Maria Farinella\\
University of Catania\\
{\tt\small gfarinella@dmi.unict.it}
}

\date{}

\maketitle
\begin{abstract}
Egocentric action anticipation is the task of predicting the future actions a camera wearer will likely perform based on past video observations.
While in a real-world system it is fundamental to output such predictions before the action begins, past works have not generally paid attention to model runtime during evaluation.
Indeed, current evaluation schemes assume that predictions can be made offline, and hence that computational resources are not limited.
In contrast, in this paper, we propose a ``streaming'' egocentric action anticipation evaluation protocol which explicitly considers model runtime for performance assessment, assuming that predictions will be available only after the current video segment is processed, which depends on the processing time of a method.
Following the proposed evaluation scheme, we benchmark different state-of-the-art approaches for egocentric action anticipation on two popular datasets. 
Our analysis shows that models with a smaller runtime tend to outperform heavier models in the considered streaming scenario, thus changing the rankings generally observed in standard offline evaluations.
Based on this observation, we propose a lightweight action anticipation model consisting in a simple feed-forward 3D CNN, which we propose to optimize using knowledge distillation techniques and a custom loss.
The results show that the proposed approach outperforms prior art in the streaming scenario, also in combination with other lightweight models.
\vfill
\end{abstract}

\section{Introduction}
Wearable devices equipped with egocentric cameras are recently attracting attention as an ideal platform to implement intelligent agents able to provide assistance to humans in a natural way~\cite{kanade2012first}.
Among the different problems studied in egocentric vision, the task of action anticipation, which consists in predicting a plausible future action before it is performed by the camera wearer, has attracted a lot of attention~\cite{camporese2020knowledge,damen2020epic,dessalene2021forecasting,furnari2020rolling,liu2020forecasting,qi2021self,sener2020temporal,wu2020learning,zhang2020egocentric}.
Indeed, from a practical point of view, being able to predict future events is fundamental when designing technologies which can assist humans in their daily and working activities~\cite{koppula2015anticipating,soran2015generating}.

\begin{figure}
    \centering
    \includegraphics[width=\linewidth]{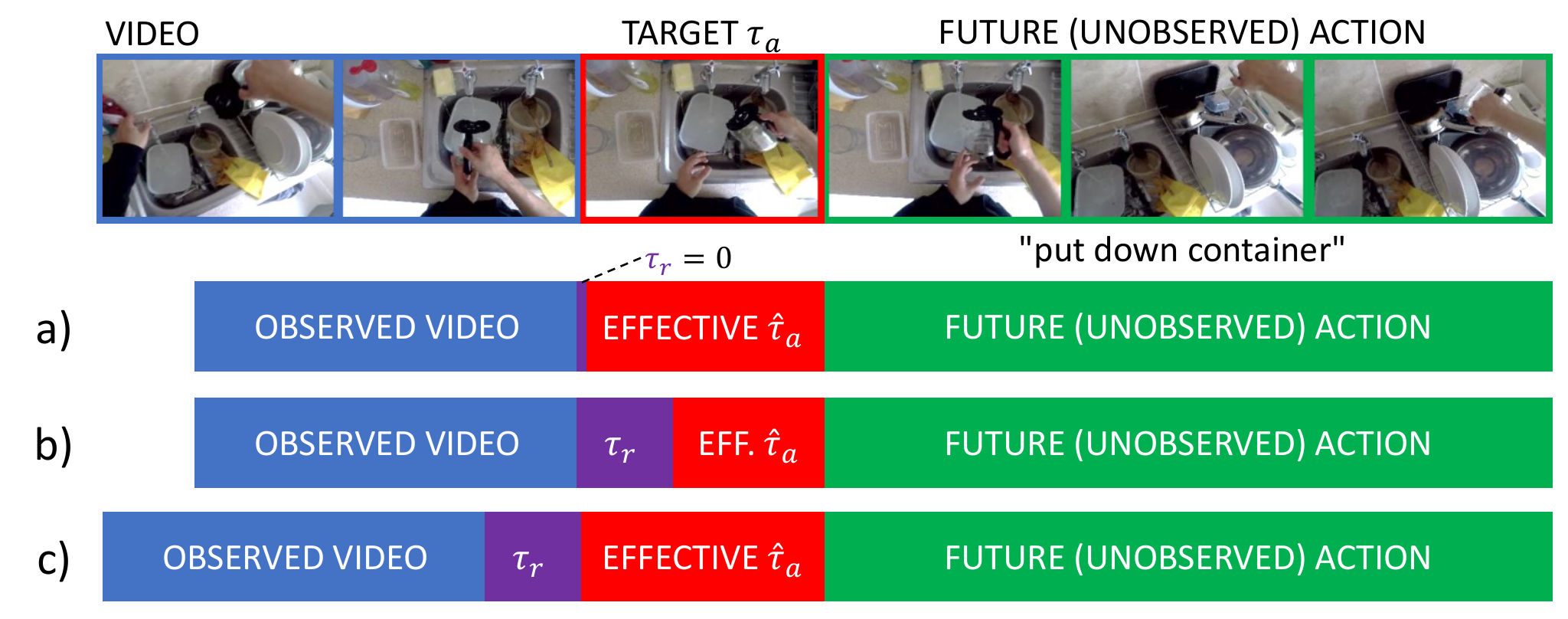}
    \caption{Different schemes to evaluate egocentric action anticipation methods. (a)
    Standard offline scenario in which runtime ($\tau_r$) is assumed to be zero. (b) Real-world case in which runtime is non-zero. The effective anticipation time ($\hat \tau_a$) is smaller than the target anticipation time ($\tau_a$).
    (c) Proposed scheme in which the observation is sampled in advance to account for the non-zero runtime. The effective anticipation time is equal to the target one.}
    \label{fig:anticipation_runtime}
\end{figure}
To make sure that action anticipation is practically useful, the predictions about the future should be available before the action is initiated by the user.
Nevertheless, previous works have not considered the computational time required to process the input video, which may lead the predictions to be late.
Actually, little attention has been paid to the effect of model runtime on performance in scenarios in which predictions need to be used in real time in general~\cite{li2020towards}, and egocentric action anticipation approaches, in particular, have been evaluated assuming a negligible runtime~\cite{damen2020epic,furnari2020rolling,sener2020temporal,liu2020forecasting,wu2020learning,zhang2020egocentric,qi2021self}.
We argue that this assumption can lead to unfair and overly optimistic evaluations and propose that egocentric action anticipation should be evaluated considering a “streaming” scenario. In these settings, the anticipation algorithm processes video segments as they become available and outputs predictions after the computational time required by the anticipation model is elapsed. 
Figure~\ref{fig:anticipation_runtime} reports different schemes to evaluate egocentric action anticipation.
In the scheme, an action anticipation model is used to predict the label of a future action (in green).
The model has been designed to anticipate actions beginning in $\tau_a$ seconds (target anticipation time) observing a video segment of $\tau_o$ seconds (observation time).
Current approaches implicitly assume that the model runtime $\tau_r$ is negligible ($\tau_r=0$) and so the prediction for a given video segment will be available right after it is passed to the model (Figure~\ref{fig:anticipation_runtime}(a)). 
However, in a realistic case in which the model runtime is larger than zero ($\tau_r>0$), predictions about future actions will be available at a later timestamp, which makes the effective anticipation time ($\hat \tau_a$) smaller than the target one ($\tau_a$) (Figure~\ref{fig:anticipation_runtime}(b)).
Since different methods are likely characterized by different runtimes, they will also have different effective anticipation times $\hat \tau_a$, which can make comparisons under the classic scheme limited.
To evaluate methods more uniformly, the temporal bounds of the observed video can be adjusted at test time to make the effective anticipation time equal to or larger than the target one: $\hat \tau_a \geq \tau_a$.
As illustrated in Figure~\ref{fig:anticipation_runtime}(c), this can be obtained by shifting the observed video backwards by $\tau_r$ seconds, hence sampling input observed videos in advance.
Under this scheme, methods with different runtimes will actually observe different video segments, which penalizes methods with a larger runtime, as we expect to happen in a real-world streaming scenario.

In this paper, we propose a new evaluation scheme which allows to asses the performance of existing egocentric action anticipation approaches in a streaming scenario by simply re-computing the timestamps at which input video segments are sampled at test time depending on the target anticipation time, the method’s estimated runtime and the length of the video observation.
Since in this scenario smaller runtimes are more effective, we propose a lightweight egocentric action anticipation model based on simple feed-forward 3D CNNs.
While feed-forward 3D CNNs are fast, we note that their performance tends to be limited as compared to full-fledged models involving different components and multi-modal observations.
We hence propose to optimize the performance of such models by using a future-to-past knowledge distillation approach which allows to transfer the knowledge from an action recognition model to the target anticipation network.
Experiments on two popular datasets, EPIC-KITCHENS-55 and EGTEA Gaze+, show that 1)~the proposed evaluation scheme induces a different ranking over state-of-the-art methods with respect to non-streaming evaluation schemes, which suggests that current evaluation protocols could be biased and incomplete, 2)~the proposed lightweight method based on 3D CNNs achieves state-of-the-art results in the streaming scenario, which advocates for the viability of knowledge distillation to make learning more data-effective for streaming action anticipation, 3)~lightweight approaches tend to outperform more sophisticated methods requiring different modalities and complex feature extractors in the streaming scenario, which suggests that more attention should be paid to runtime optimization when tackling anticipation tasks.

In sum, the contributions of this work are as follows: 1) We propose a streaming evaluation scheme to assess egocentric action anticipation methods fairly and practically; 2) We benchmark several state-of-the-art methods and show that the proposed streaming evaluation scheme induces a different ranking with respect to offline evaluation, indicating the limits of current performance assessment protocols; 3) We propose a lightweight action anticipation model based on feed-forward 3D CNNs and optimized through knowledge distillation. Experiments show that the proposed approach obtains state-of-the-art performance in the streaming scenario, achieving results which are competitive with respect to heavier approaches at a fraction of their computational cost.

\section{Related Work}
\label{sec:related}

\paragraph{Efficient, Online and Streaming Vision}
Works in image and video understanding generally assumed that computation can be performed offline~\cite{he2016deep,simonyan2014very,szegedy2015going,rastegari2016xnor,he2015spatial,lin2017focal,ren2016faster,carreira2017quo,feichtenhofer2019slowfast,he2015spatial,furnari2020rolling,sener2020temporal,liu2020forecasting}.
Hence, little attention has been paid by these methods to model runtime.
Another line of research has devoted its attention to the development of computationally efficient models for image recognition~\cite{tan2019efficientnet,howard2017mobilenets,iandola2016squeezenet}, object detection~\cite{liu2016ssd,redmon2016you,redmon2017yolo9000} and action recognition~\cite{feichtenhofer2020x3d}. 
These works have usually been evaluated considering standard evaluation schemes, as well as reporting model runtime or number of FLoating point Operations Per Second (FLOPS).
Previous works have also tackled the problem of online video processing, studying online action detection~\cite{de2016online}, early action recognition~\cite{hoai2014max,sadegh2017encouraging}, action anticipation~\cite{damen2020epic,gao2017red,furnari2020rolling} and next active object prediction~\cite{furnari2017next}.
These tasks require the developed algorithm to make predictions before a video event is completely or even partially observed.
However, they generally do not evaluate models according to runtime, hence assuming computational resources not to be finite.
Some works have recently considered a ``streaming'' scenario in which algorithms should be evaluated considering the time in which the predictions are available depending on model runtime~\cite{kristan2017visual,li2020towards}.

We build on works on efficient computer vision~\cite{feichtenhofer2020x3d,howard2017mobilenets}, online video processing~\cite{de2016online} and streaming perception~\cite{li2020towards}.
However, differently from these works, we study the streaming scenario within the task of egocentric action anticipation, in which the timeliness of predictions is fundamental to ensure their practical utility. 
To this aim, we propose an evaluation scheme which explicitly takes into account model runtime and the streaming nature of video, allowing to compare algorithms in a more homogeneous and practical way.

\paragraph{Egocentric Action Anticipation}
Different works have tackled this task~\cite{furnari2020rolling,sener2020temporal,dessalene2021forecasting,liu2020forecasting}.
Previous approaches have considered baselines designed for action recognition~\cite{damen2020epic}, defined custom losses~\cite{furnari2018leveraging}, modeled the evolution of scene attributes and action over time~\cite{miech2019leveraging}, disentangled the tasks of encoding and anticipation~\cite{furnari2020rolling}, aggregated features over time~\cite{sener2020temporal}, predicted motor attention~\cite{liu2020forecasting}, leveraged contact representations~\cite{dessalene2021forecasting}, mimicked intuitive and analytical thinking~\cite{zhang2020egocentric}, and predicted future representations~\cite{wu2020learning}.
While these approaches have been designed to maximize performance when predicting the future, they have never been evaluated in a streaming scenario. In fact model runtime has not been reported or even mentioned in past works.

We benchmark some representatives of these methods and show experimentally that their performance is limited in the streaming scenario.
We contribute a novel streaming evaluation scheme which can be used to assess performance considering a more realistic and practical scenario and propose a lightweight model for egocentric action anticipation which can be used in resource-constrained settings.

\paragraph{Knowledge Distillation}
Knowledge distillation techniques aim to transfer the knowledge from a large and expensive neural network (the teacher) to a small and lightweight network (the student).
Initial approaches used the logits of the teacher model as a supervisory signal for the student~\cite{hinton2015distilling}.
These methods aimed to minimize the divergence between the probability distributions predicted by the student and the teacher for a given input example, with the goal of providing a richer learning objective as compared to deterministic ground truth labels.
Other approaches used the activations of the intermediate layers of the teacher to guide the optimization of the student~\cite{Huang2017,romero2014fitnets}, whereas
some methods modeled the relationships between the outputs of different layers of the network to provide guidance to the student on how to process the input example~\cite{yim2017gift,passalis2020heterogeneous}. 

We use knowledge distillation to optimize the performance of a lightweight action anticipation model and achieve competitive results in the streaming scenario. 
Differently from the aforementioned works, we do not transfer knowledge from a complex model to a simple one. Instead, we use an action recognition model looking at a future action as the teacher, and a lightweight model looking at a past video snippet as the student. 
This distillation approach is referred to in this paper as ``future-to-past''. 
We propose a loss which facilitates knowledge transfer using the teacher to instruct the student on which spatiotemporal features are more discriminative for action anticipation.

\paragraph{Knowledge Distillation for Future Prediction}
Few previous works used knowledge distillation to improve action anticipation.
Some works used knowledge distillation implicitly by encouraging the model to predict representations of future frames, which were later used to make the predictions~\cite{vondrick2016anticipating,gao2017red,wu2020learning}.
Other works have explicitly considered distillation approaches to transfer knowledge from a fixed teacher with label smoothing~\cite{camporese2020knowledge} or from an action recognition model~\cite{tran2019back,Fernando21}.
Other works applied knowledge techniques to early action prediction~\cite{Wang2019}.

Similarly to these works, we study methods to transfer knowledge from an action recognition model to the target anticipation network.
However, differently from previous works, our main objective is to assist the optimization of a lightweight and computationally efficient model.

\section{Streaming Evaluation Scheme}
\label{sec:streaming_evaluation}

\begin{figure}
    \centering
    \includegraphics[width=\linewidth]{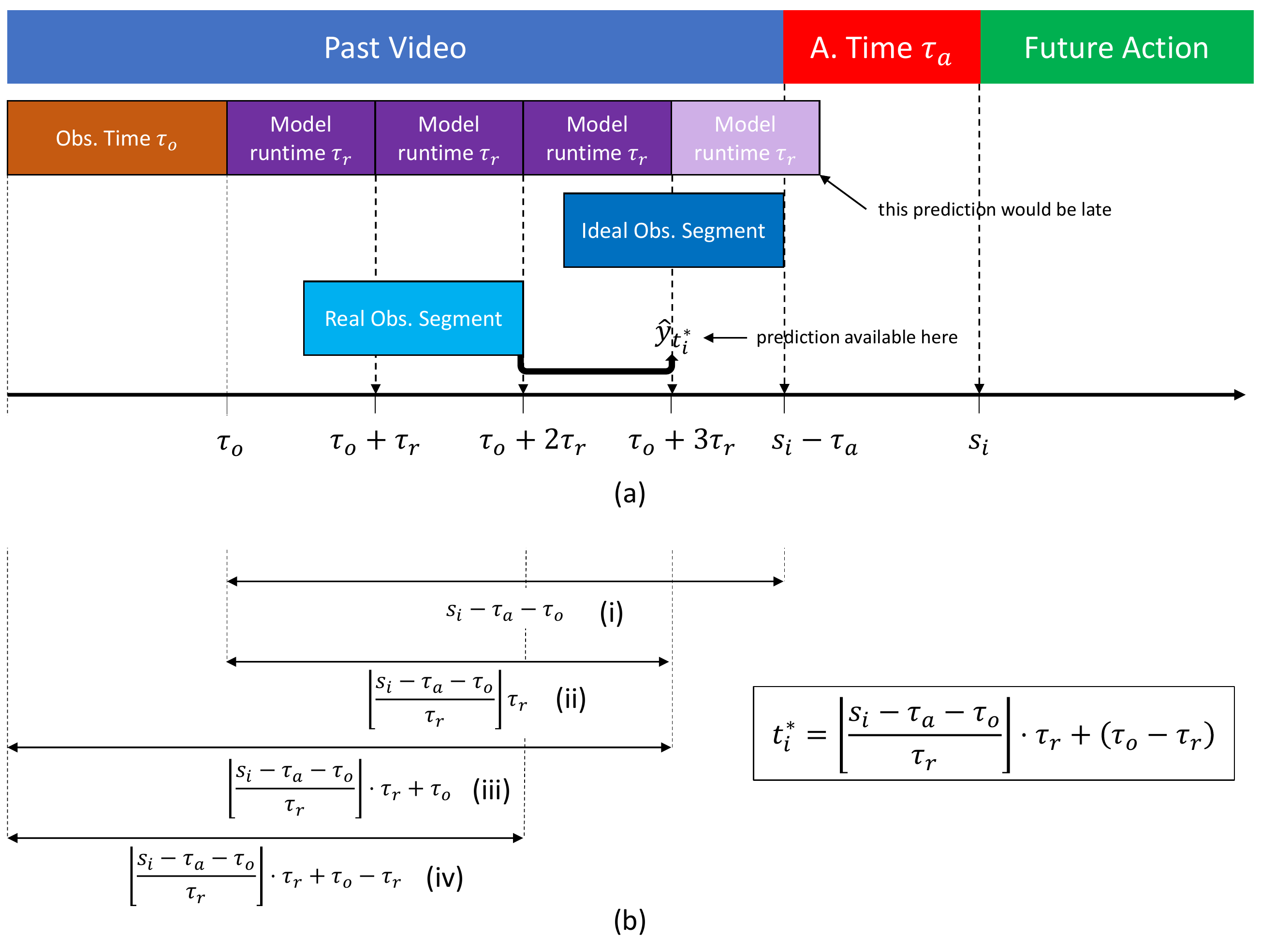}
    \caption{Video sampling induced by the proposed streaming evaluation protocol.
    (a) Due to non-zero observation time and runtime, models output predictions only at given timestamps. Hence the real observed segment is different form the ideal one. (b) Scheme of Eq.~(\ref{eq:quantization}). See text for discussion.}
    \label{fig:quantization}
\end{figure}

Let $V$ be the input video, and let $V_{t_s:t_e}$ denote a video clip starting at timestamp $t_s$ and ending at timestamp $t_e$. Let $\phi$ be the model designed to process videos of length $\tau_o$ (observation time) and anticipate actions happening after $\tau_a$ seconds (anticipation time). 
At a given timestamp $t$, the algorithm processes the most recent video segment $V_{t-\tau_o:t}$.
Let $\tau_r$ be the time required by the model to process a video segment and output a prediction (model runtime).
The prediction $\phi(V_{t-\tau_o:t})$ will be available at timestamp $t+\tau_r$. We hence denote it as $\hat y_{t+\tau_r}=\phi(V_{t-\tau_o:t})$. 
Similarly, we denote a prediction available at time $t$ as $\hat y_t =\phi(V_{t-\tau_o-\tau_r:t-\tau_r})$. This notation makes explicit that, in the presence of a large runtime $\tau_r$, the input video segment should be sampled ahead of time to preserve a correct anticipation time. 
We assume that resources are limited and only a single GPU process is allowed at a time. This is a realistic scenario when algorithms are deployed to a mobile device such as smart glasses. Note that, in this case, the runtime will not be negligible ($\tau_r>0$), and hence the model will not be able to make predictions at every single frame of the video. 
Specifically, predictions will be available only at selected timestamps $\tau_o + k \cdot \tau_r,k \in N^+$. 
This happens because the model first needs to wait for $\tau_o$ seconds to fill the video buffer, then it has to wait for the runtime $\tau_r$ to make the next prediction.
As a consequence, there will be a difference between the ideal video segment in offline settings and the one actually sampled when processing the video in streaming mode.
Figure~\ref{fig:quantization}(a) illustrates an example of the new video sampling introduced by the proposed streaming action anticipation scenario and the related difference between ideal and real observed video segments.

For consistency with past benchmarks and to use the labels available in current action anticipation datasets, we evaluate anticipation methods only at specific timestamps sampled $\tau_a$ seconds before the beginning of each action.
In particular, let $A_i=(s_i,e_i,y_i,)$ be the $i^{th}$ labeled action of a test video $V$, where $s_i$ denotes the action start timestamp, $e_i$ denotes the action end timestamp, and $y_i$ denotes the action label. 
The action $A_i$ will be associated to the most recent prediction made $\tau_a$ seconds before the beginning of the action, at timestamp $s_i-\tau_a$. It is worth noting that a prediction might not be available exactly at the required timestamp (see Figure~\ref{fig:quantization}(a)), so we will consider the most recent timestamp at which a prediction is available, which is given by the following formula:
\begin{equation}
t_i^*=\left \lfloor \frac{s_i-\tau_a-\tau_o}{\tau_r}\right \rfloor \cdot \tau_r+\tau_o-\tau_r.
\label{eq:quantization}
\end{equation}
Figure~\ref{fig:quantization}(b) reports a scheme of the formula. The term
$\left \lfloor \frac{s_i-\tau_a-\tau_o}{\tau_r}\right \rfloor$ computes the number of time slots of length $\tau_r$ completely included in the segment of length $s_i-\tau_a-\tau_o$ ((i) in Figure~\ref{fig:quantization}(b)). Subtracting $\tau_o$ is necessary because the first video segment will be processed only after $\tau_o$ seconds. The product $\left \lfloor \frac{s_i-\tau_a-\tau_o}{\tau_r}\right \rfloor \cdot \tau_r$ ((ii) in Figure~\ref{fig:quantization}(b)) quantizes the timestamp. The term $ + \tau_o$ adds the observation time which had been previously subtracted ((iii) in Figure~\ref{fig:quantization}(b)) and the term $-\tau_r$ subtracts the runtime needed to obtain the prediction ((iv) in Figure~\ref{fig:quantization}(b)).
We will hence associate the following prediction to action $A_i$: $\hat y_{t_i^*}=\phi(V_{t_i^*-\tau_o:t_i^*})$.

It is worth noting that, the model runtime $\tau_r$ is estimated, the proposed evaluation scheme can be easily implemented simply reworking the evaluation timestamps using Equation~(\ref{eq:quantization}).
We would also like to note that this evaluation scheme is not limited to egocentric action anticipation, but could be applied in third-person anticipation scenarios as well.

\paragraph{Evaluation Measures}
In principle, the proposed evaluation scheme can be implemented with any performance measure for egocentric action anticipation.
In this work, we consider Mean Top-5 Recall (MT5R)~\cite{furnari2018leveraging}.
Similarly to Top-5 accuracy, Mean Top-5 Recall considers a prediction correct if the ground truth action is included in the top-$5$ predictions.
Differently from Top-5 accuracy, Mean Top-5 Recall is a class-aware metric in which performance indicators obtained for each class are averaged to obtain the final score.
As a result, Mean Top-5 Recall is more robust to the class imbalance usually contained in large scale datasets with a long tail distribution~\cite{damen2020epic}.

\section{Method}
\label{sec:method}
Based on the observation that model runtime influences performance in the considered streaming evaluation scheme, we propose a lightweight egocentric action anticipation approach based on simple feed-forward 3D CNNs.
Differently from full-fledged anticipation models involving multi-modal feature extraction and sequence processing~\cite{furnari2020rolling,sener2020temporal,gao2017red}, feed-forward 3D CNNs are simple and fast, but also harder to optimize for action anticipation. To tackle this issue, we propose a training scheme based on knowledge distillation which we show to greatly improve performance of feed-forward 3D CNNs.

\begin{figure}
    \centering
    \includegraphics[width=\linewidth]{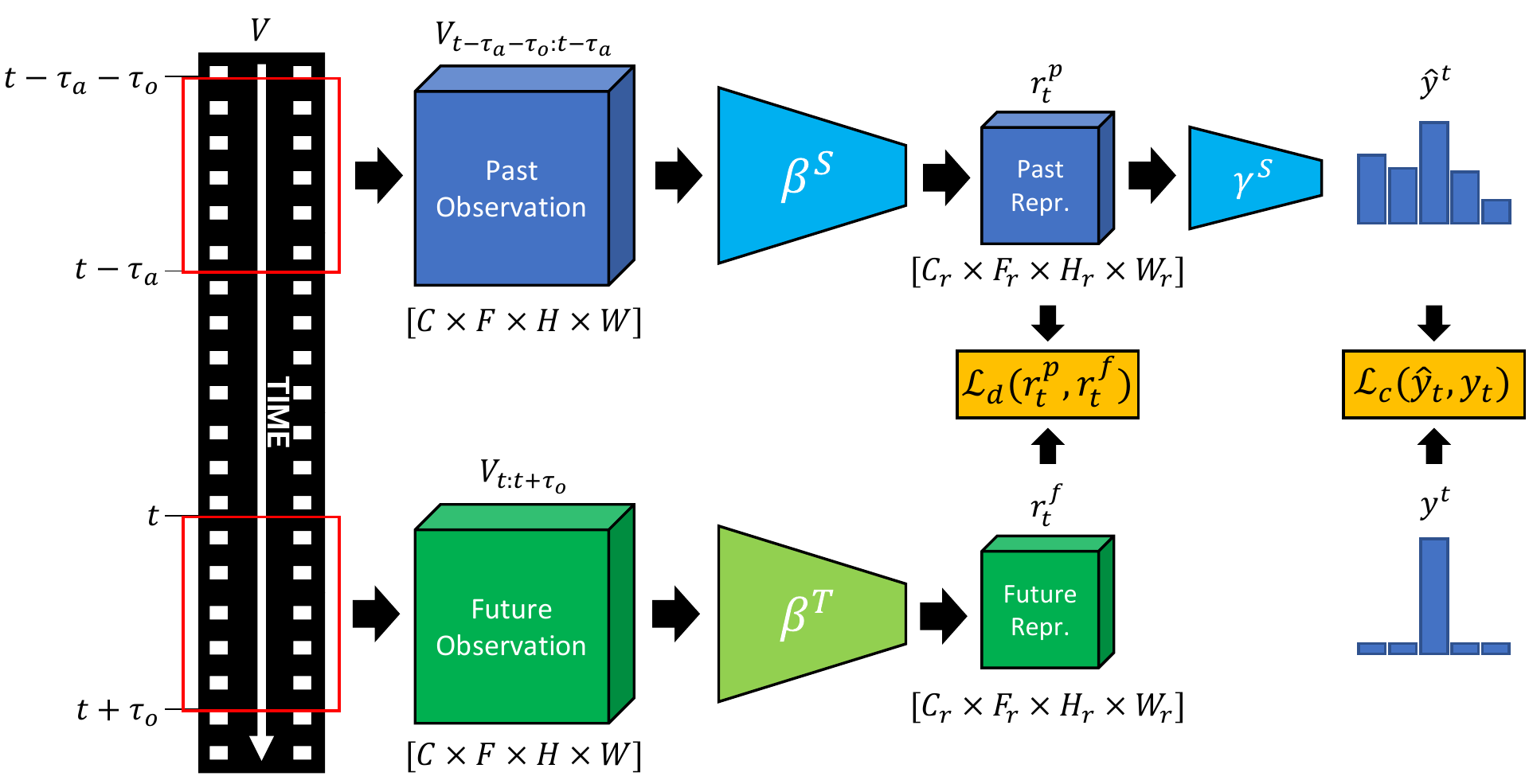}
    \caption{Proposed training scheme. The teacher (bottom) processes future video segments $V_{t:t+\tau_o}$, while the student (top) processes past video segments $V_{t-\tau_a-\tau_o:t-\tau_a}$. We use a classification loss ($\mathcal{L}_c$) to enforce that the label predicted by the student is correct and a distillation loss ($\mathcal{L}_d$) to encourage consistency between the representations extracted by student and teacher.}
    \label{fig:training}
\end{figure}

\paragraph{Proposed Training Scheme}
Figure~\ref{fig:training} illustrates the proposed training procedure, which operates over both labeled and unlabeled examples. 
Given a video $V$ and a timestamp $t$, we form a training example considering a pair of videos including a past observation $V_{t-\tau_a-\tau_o:t-\tau_a}$, a future observation $V_{t:t+\tau_o}$, and a label $y_t$ denoting the class of the action included in the future observation $V_{t:t+\tau_o}$. 
Depending on the sampled timestamp $t$, the future video segment might not be associated to any action, in which case we will say that the example is unlabeled denote its label with $y_t=\varnothing$.
As a teacher, we use a 3D CNN $\phi^T$ pre-trained to perform action recognition from input videos of resolution $C \times F \times H \times W$, where $C$ is the number of channels ($C=3$ for RGB videos), $F$ is the number of frames, $H$ and $W$ are the video frame height and width respectively. The teacher $\phi^T=\beta^T \circ \gamma^T$ is composed of a backbone $\beta^T$ which extracts spatio-temporal representations of resolution $C_r \times F_r \times H_r \times W_r$, as well as a classifier $\gamma^T$ which predicts a probability distribution over classes.
The student model $\phi^S=\beta^S \circ \gamma^S$ has the same structure as the teacher $\phi^T$ and it is initialized with the same weights. 
We train the model by feeding past observations to the student and future observations to the teacher.
At each training iteration, we extract the internal representation of the past segment $r_t^p=\beta^S(V_{t-\tau_a-\tau_o:t-\tau_a })$, the representation of the paired future segment $r_t^f=\beta^T(V_{t:t+\tau_o })$, and the predicted future action label $\hat y_t = \gamma^S(r_t^p)$.

We hence train the student to both classify the past observation correctly and extract representations coherent with the ones of the future segment using the following loss:
\begin{equation}
\label{eq:loss}
    \mathcal{L}=\lambda_d \mathcal{L}_d (r_t^p, r_t^f) + [y_t \neq \varnothing] \lambda_c \mathcal{L}_c (\hat y_t, y_t),
\end{equation}
where $\mathcal{L}_d$ is a distillation loss used to encourage the representations of the past and future segments to be coherent, $\mathcal{L}_c$ is a classification loss which aims to reduce the classification error of the student (e.g., cross entropy loss), $\lambda_d$ and $\lambda_c$ are hyper-parameters used to regulate the contributions of the two losses, and the Iverson bracket term $[y_t\neq \varnothing]$ is used to avoid computing the classification loss when the example is unlabeled.

\begin{figure}
    \centering
    \includegraphics[width=\linewidth]{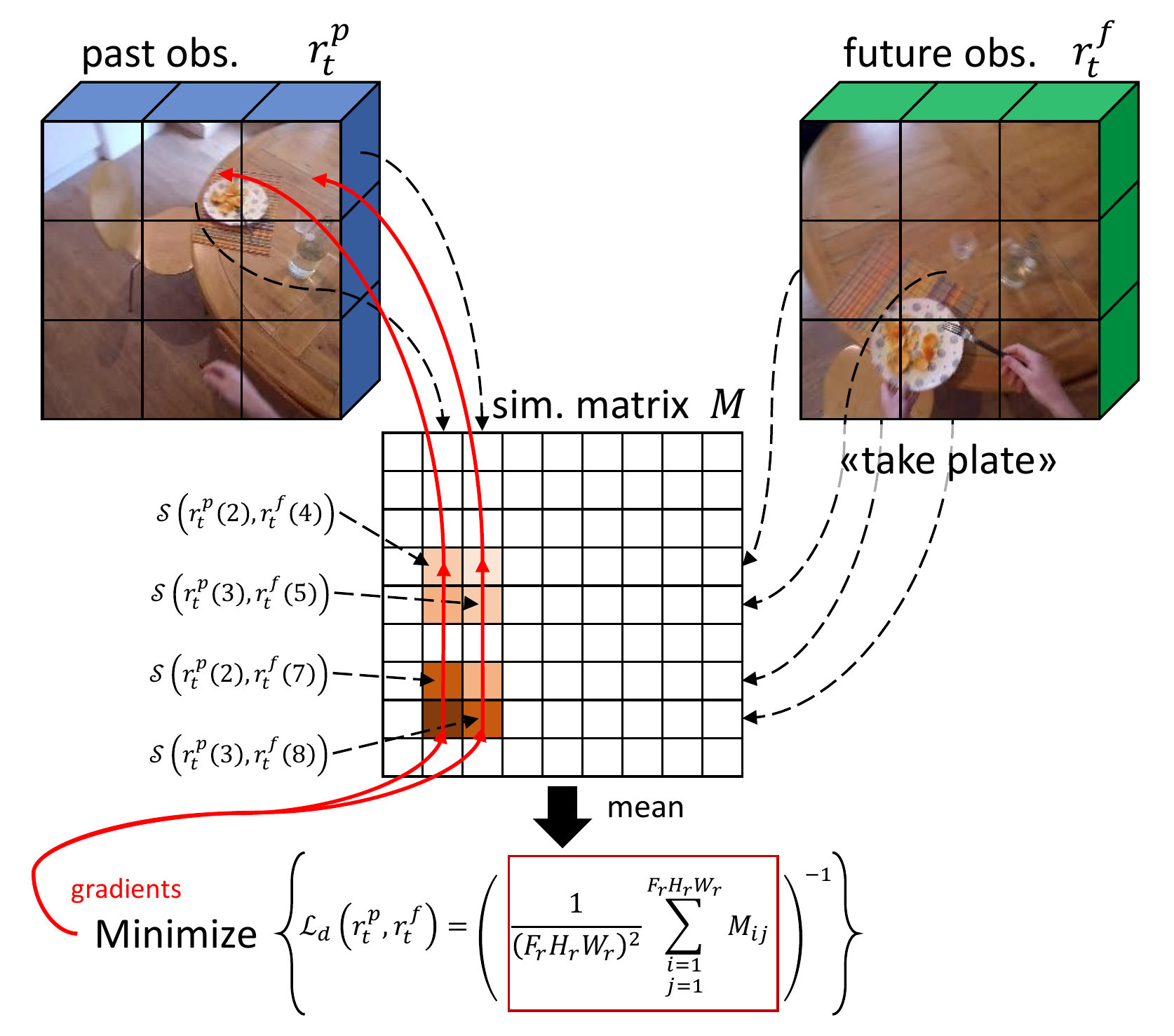}
    \caption{An illustration of the proposed distillation for a ``take plate'' future action.
    Minimizing the loss maximizes the similarity between all pairs of past and future representations, which encourages the student to extract ``plate'' features. See text for details.}
    \label{fig:loss}
\end{figure}

\paragraph{Proposed Knowledge Distillation Loss}
Unlike from classic knowledge distillation, in our training protocol, the teacher and student models process different but related inputs (i.e., the past and future video segments).
Let $r_t^p (i)$ and $r_t^f (i)$ be the $C_r$-dimensional representations corresponding to the $i^{th}$ spatiotemporal locations of the $r_t^p$ and $r_t^f$ tensors.
One way to enforce knowledge distillation would be to maximize the similarity between corresponding representations $r_t^p (i)$ and $r_t^f (i)$ directly, e.g., using the Mean Squared Error (MSE) loss.
However, since the inputs of the two networks are different, we expect their representations to be spatio-temporally misaligned.
To mitigate this misalignment, we propose to maximize similarity between all $(r_t^p (i),r_t^f (j))$ pairs $\forall i,j\in{1,\ldots,F_r \cdot H_r \cdot W_r}$.
Practically, we define a future-to-past similarity matrix $M$ as follows:
\begin{equation}
    M_{ij} = \frac{r_t^p(i) \cdot r_t^f(j)}{||r_t^p(i)||_2 \cdot ||r_t^f(j)||_2}.
\end{equation}
Note that the general term of the matrix $M_{ij}$ is the cosine similarity between the two representations $r_t^p (i)$ and $r_t^f (j)$.
We hence maximize the values of $M$ by minimizing the following loss:
\begin{equation}
    \mathcal{L}_d(r_t^p,r_t^f) = \left( \frac{1}{(F_r H_r W_r)^2} \sum_{\substack{i=1\\j=1}}^{F_r H_r  W_r} M_{ij} \right)^{-1}.
\end{equation}
The main rationale behind this training objective is to encourage the student network to pay attention to spatio-temporal locations in the past observation which contain semantic content important for action recognition in the future, even if these are not spatio-temporally aligned. 
For example, if the future action is “take plate”, the student network should learn to extract “plate” features from the past segment even if the features occur at different spatio-temporal regions. Figure~\ref{fig:loss} illustrates the proposed loss.

\section{Experimental Settings}
\label{sec:experimental_settings}
\paragraph{Datasets}
We perform experiments on two datasets: EPIC-KITCHENS-55~\cite{damen2020epic} and EGTEA Gaze+~\cite{li2021eye}.

EPIC-KITCHENS-55~\cite{damen2020epic} is composed of $432$ videos acquired by $32$ subjects and labeled with $39,595$ action segments with a taxonomy of $125$ verbs and $352$ nouns. 
We split the public dataset following~\cite{furnari2020rolling} to obtain training set of $232$ videos and $23,493$ action segments, and a validation set of $40$ videos and $4,979$ segments. We report results on the validation set.
We consider all unique $(verb, noun)$ pairs appearing in the public set to obtain $2,513$ distinct action classes.

EGTEA Gaze+~\cite{li2021eye} includes $86$ videos acquired by $28$ different subjects, labeled with $10,325$ action segments including $19$ verbs, $51$ nouns and $106$ action classes.
We randomly split the dataset into a training set containing $65$ videos and a test set containing $21$ videos. We report results on the test set.

\paragraph{Compared Methods}
We benchmark different egocentric action anticipation approaches using to both a classic offline evaluation scheme and the proposed streaming protocol described in Section~\ref{sec:streaming_evaluation}.
We choose different approaches spanning from full-fledged but computationally expensive methods to the lightweight baselines:
\begin{itemize}
    \item \textbf{Full-fledged methods}: We consider the Rolling-Unrolling LSTMs (RULSTM) proposed in~\cite{furnari2020rolling}, and an Encoder-Decoder (ED) architecture based on~\cite{gao2017red}. These approaches make use of LSTMs and consider different input modalities including representations extracted from RGB frames, optical flow, and object-based features. Due to the multi-modal data extraction and processing, these approaches are accurate but computationally expensive.
    \item \textbf{Action recognition baselines}: We consider methods achieving state-of-the-art results in action recognition. 
    Specifically, we include different 3D feed-forward CNNs of varying computational complexity: a computationally expensive I3D~\cite{carreira2017quo} network processing clips of resolution $3 \times 64 \times 224 \times 224$, the more efficient SlowFast~\cite{feichtenhofer2019slowfast} network based of ResNet50 with inputs of resolution $3 \times 32 \times 224 \times 224$, the computationally optimized X3D-XS architecture~\cite{feichtenhofer2020x3d} processing inputs of size $3 \times 4 \times 160 \times 160$, and the lightweight R(2+1)D~\cite{tran2018closer} CNN based on ResNet18 and processing videos of resolution $3 \times 16 \times 64 \times 64$. 
    \item \textbf{Lightweight baselines}: We also include two baselines designed to be very lightweight, but likely to be less accurate. In particular, we consider Temporal Segment Networks (TSN)~\cite{wang2016temporal} and a simple LSTM~\cite{gers1999learning} which takes as input representations of the RGB frames obtained using a BNInception 2D CNN pre-trained for action recognition.
\end{itemize}

\paragraph{Implementation Details}
We estimate all model runtimes on a NVIDIA K80 GPU assuming that resources would be limited at run time in a real scenario.
As in previous works~\cite{furnari2020rolling}, models are trained to predict a probability distribution over actions ($(verb, noun)$ action pairs), while verb and noun probability distributions are obtained by marginalization.
We instantiate our method using an R(2+1)D backbone~\cite{tran2018closer} based on ReseNet18~\cite{he2016deep} as implemented in PyTorch~\cite{pytorch} which processes video clips of resolution $3 \times 16 \times 64 \times 64$.
We choose this model for its good trade-off between performance and computational efficiency.
Video clips are sampled with a temporal stride of $2$ frames.
We first train the teacher for action recognition, then initialize the student with the teacher's weights and optimize only the student network with the proposed training scheme. 
See Appendix~\ref{appendix:implementation} for more details.

\section{Results}
\label{sec:results}

\subsection{Proposed Streaming Benchmark}
\label{sec:benchmark}
In Table~\ref{tab:ek55_results} and Table\ref{tab:egtea_results} we report the performances of the proposed methods on the considered datasets. Results are reported using both a classic offline evaluation protocol (columns $4-6$) and the proposed streaming evaluation scheme (columns $7-9$).
Column $2$ of Table~\ref{tab:ek55_results} also reports the runtime in milliseconds (R.TIME) whereas column $3$ reports the framerate if frames per second (FPS).
Verb, noun and action (ACT.) performance is reported using Mean Top-5 Recall\%.
Best results per column are reported in bold, whereas second-best results are underlined.

\paragraph{EPIC-KITCHENS-55}
\begin{table}[t]
 \centering
 \adjustbox{max width=\linewidth}{
 \setlength{\tabcolsep}{2.5pt}
\renewcommand{\arraystretch}{1.2}
 \begin{tabular}{lrr|rrr|rrr}
 \multicolumn{3}{}{} & \multicolumn{3}{c}{\textbf{OFFLINE}} & \multicolumn{3}{c}{\textbf{STREAMING}} \\
   \hline
 \textbf{METHOD} & \textbf{R.TIME} & \textbf{FPS} & \textbf{VERB} & \textbf{NOUN} & \textbf{ACT}. & \textbf{VERB} & \textbf{NOUN} & \textbf{ACT.} \\ 
   \hline
   RULSTM\cite{furnari2020rolling}  &  724.98 & 1.38 & 21.91 & \textbf{27.19} & \textbf{14.81} & 19.80 & 24.04 & 11.40\\ 
   ED\cite{gao2017red} & 100.56 & 9.94 & 20.97 & 23.35 & 10.60  & 19.87 & 22.41 & 9.60 \\ 
   \hline
   I3D\cite{carreira2017quo} &  275.26 & 3.63 & 21.50 & 23.68 & 11.77& 22.08 & 22.63 & 10.83\\ 
   SlowFast\cite{feichtenhofer2019slowfast} & 173.73 & 5.76 & 19.97 & 21.10 & 9.87 & 18.53 & 20.96 & 9.31 \\
   X3D-XS\cite{feichtenhofer2020x3d} &  142.50 & 7.02 & 17.49 & 19.47 & 8.54 & 17.49 & 19.47 & 8.54 \\
   R(2+1)D\cite{tran2018closer} &  41.41 & 24.15 & 15.31 & 17.09 & 8.10 & 14.52 & 16.82 & 7.89\\ 
   \hline
   LSTM\cite{gers1999learning} & 25.96 & 38.52 & 21.59 & 24.62 & 12.35 & 21.22 & \underline{24.80} & 12.18 \\
   TSN\cite{li2020towards} & 19.20 & 52.08 & 19.76 & 21.55 & 10.28& 19.80 & 21.47 & 10.25\\ 
   \hline
   \hline
   DIST-R(2+1)D &  41.41 &  24.15 & \textbf{23.67} & 24.43 & 12.83& \textbf{23.41} & 23.83 & \underline{12.31} \\
  LSTM+DIST-R(2+1)D & 67.37 & 14.84 & \underline{23.51} & \underline{26.11} & \underline{14.35} & \underline{22.65} & \textbf{26.60} & \textbf{13.55} \\
  \hline
\end{tabular}}
\caption{Results on EPIC-KITCHENS-55 in Mean Top-5 Recall\%}
 \label{tab:ek55_results}
\end{table}

The results reported in Table~\ref{tab:ek55_results} highlights how methods with large runtimes are more optimized for performance and tend to outperform competitors in the standard offline evaluation scenario. 
For example, ED is surpassed by RULSTM, which is also based on LSTMs but makes use of object-based features, which have a significant impact on runtime (ED has a runtime of $100.56 ms$, whereas RULSTM has a runtime of $724.98 ms$). 
Likewise, I3D outperforms SlowFast and X3D-XS probably due  to the larger input size, at the cost of a larger runtime ($275.26 ms$ of I3D vs $173.73 ms$ of SlowFast and $142.5 ms$ of X3D XS).
R(2+1)D is the lightest method among the considered 3D CNNs, with a framerate of $24.15 fps$, but achieves limited performance ($8.10$ offline action performance).
In general, methods based on simple 3D CNNs tend to perform worse than highly optimized methods such as RULSTM.
Notably, the proposed approach based on knowledge distillation improves the offline action performance of the R(2+1)D baseline by $+4.73$ ($12.83$ of DIST-R(2+1)D vs $8.10$ of R(2+1)D-M)

It can be noted that runtime significantly affects performance when the streaming evaluation scenario is considered.
Indeed, by comparing columns $6$ and $9$, it is clear that, while the performances of all methods decrease when passing from offline to streaming evaluation (compare columns $6$ and $9$), methods characterized by smaller runtimes are characterized by a smaller performance gap.
For example, RULSTM has an offline action performance of $14.81$, which is reduced to $11.40$ in the streaming scenario ($-3.41$), whereas TSN, which is the fastest approach, has an offline action performance of $10.28$ which is very comparable to the performance of $10.25$ achieved in the streaming scenario (only $-0.03$).
Interestingly, the performance of the lightweight LSTM approach (runtime of $25.96ms$) is barely affected when passing from an offline to an online evaluation scheme ($12.35$ offline action performance vs $12.18$ streaming action performance).

The last row of Table~\ref{tab:ek55_results} reports the performance of a method obtained by combining the LSTM with the proposed model based on knowledge distillation.
Results point out that the obtained LSTM+DIST-R(2+1)D, achieves good results at a small computational cost.
Indeed, LSTM+DIST+R(2+1)D achieves a similar streaming action performance of $13.55$, which significantly outperform other approaches such as RULSTM in the streaming scenario while running much faster with a runtime of $67.37ms$ and a framerate of $14.84fps$.

\begin{table}[t]
\centering
\adjustbox{max width=\linewidth}{
\setlength{\tabcolsep}{2.5pt}
\renewcommand{\arraystretch}{1.2}
\begin{tabular}{lrr|rrr|rrr}
 \multicolumn{3}{}{} & \multicolumn{3}{c}{\textbf{OFFLINE}} & \multicolumn{3}{c}{\textbf{STREAMING}} \\
   \hline
 \textbf{METHOD} & \textbf{R.TIME} & \textbf{FPS} & \textbf{VERB} & \textbf{NOUN} & \textbf{ACT}. & \textbf{VERB} & \textbf{NOUN} & \textbf{ACT.} \\ 
  \hline
  RULSTM\cite{furnari2020rolling} & 724.98 & 1.38 & \underline{78.53} & \textbf{71.11} & \textbf{59.79} & 71.17 & 60.12 & 49.17 \\ 
  ED\cite{gao2017red} & 100.56 & 9.94 & 73.54 & 64.99 & 52.78 & 73.79 & 62.68 & 52.23 \\
  \hline
  I3D\cite{carreira2017quo} & 275.26 & 3.63 & 77.37 & 65.59 & 52.53 & 70.12 & 58.95 & 47.06 \\ 
  SlowFast\cite{feichtenhofer2019slowfast} & 173.73 & 5.76 & 68.08 & 58.56 & 44.16 & 66.45 & 55.74 & 42.45 \\
  X3D-XS\cite{feichtenhofer2020x3d} & 142.50 & 7.02 & 70.79 & 61.73 & 45.16 & 68.42 & 58.81 & 44.93 \\
  R(2+1)D\cite{tran2018closer} & 41.41 & 24.15 & 68.38 & 56.83 & 42.52 & 67.52 & 54.61 & 41.48 \\
  \hline
  LSTM\cite{gers1999learning} & 25.96 & 38.52 &  77.80 & 67.68 & \underline{58.15} & \underline{77.16} & \underline{67.02} & \underline{56.52} \\
  TSN\cite{li2020towards} & 19.20 & 52.08 & 69.86 & 58.89 & 46.49 & 70.01 & 58.58 & 45.99\\ 
  \hline
  \hline
  DIST-R(2+1)D & 41.41 & 24.15 & 76.29 & 64.84 & 49.82 & 73.80 & 65.51 & 49.87 \\
  LSTM+DIST-R(2+1)D & 67.37 & 14.84 & \textbf{80.75} & \underline{70.45} & 57.95 & \textbf{79.48} & \textbf{69.79} & \textbf{57.33}\\
  \hline
\end{tabular}}
\caption{Results on EGTEA Gaze+ in Mean Top-5 Recall\%}
\label{tab:egtea_results}
\end{table}

It is worth noting that the offline and streaming evaluation schemes tend to induce different rankings of the methods, which suggests that classic offline evaluation offers only a limited picture when algorithms need to be deployed to real hardware with limited resources.
Notably, the best performing approach in the streaming scenario is the proposed model based on R(2+1)D networks and knowledge distillation, also in combination with the LSTM model.

\paragraph{EGTEA Gaze+}
Coherently with previous findings, the results reported in Table~\ref{tab:egtea_results} highlight that more optimized methods such as RULSTM achieve better offline performance at the cost of larger runtimes, whereas, when passing to the streaming scenario, lightweight methods tend to have an advantage. 
As an example, the LSTM achieves an offline action performance of $58.15$, which is lower ($-1.64$) as compared to the offline performance of RULSTM ($59.79$).
However, in the streaming scenario, the LSTM scores an action performance of $56.52$, which is much higher ($+7.35$) than the action performance of RULSTM ($49.17$).
Also on this dataset, the proposed distillation approach improves the performance of R(2+1)D-based methods.
Indeed, DIST-R(2+1)D improves the offline action performance of R(2+1)D by $+7.3$.
The best performance in the streaming scenario is obtained also in this case by LSTM+DIST-R(2+1)D, which obtains a streaming action performance of $57.33$ with a runtime of $67.67ms$ and a framerate of $14.84fps$.

\subsection{Ablation Study}
\label{sec:ablation}
To analyze the effect of the different components of the proposed model, in this Section, we report the results of ablation experiments conducted on EPIC-KITCHENS-55.

\begin{table}[t]
\centering
\adjustbox{max width=0.9\linewidth}{
\setlength{\tabcolsep}{2.5pt}
\renewcommand{\arraystretch}{1.2}
\begin{tabular}{lllrrr}
  \hline
\textbf{Data} & \textbf{Examples} & \textbf{Dist.} & \textbf{VERB} & \textbf{NOUN} & \textbf{ACT.} \\ 
  \hline
  Supervised & $23,493$ & No & 15.31 & 17.09 & 8.10 \\ 
  Supervised & $23,493$ & Yes & 20.26 & 22.40 & 10.55 \\
  \hline
  Augmented & $993,899$ & No & 17.91 & 21.77 & 9.35 \\ 
  Augmented & $993,899$ & Yes & \underline{21.82} & \underline{23.71} & \underline{10.66} \\ 
  \hline
  All & $3,584,241$ & Yes & \textbf{23.67} & \textbf{24.43} & \textbf{12.83} \\ 
  \hline
\end{tabular}}
\caption{Effect of Data and Knowledge Distillation} 
\label{tab:ablation_data}
\end{table}

\paragraph{Effects of Knowledge Distillation on Learning}
We observe that the considered future-to-past knowledge distillation scheme regularizes learning in two ways:
1) It guides the anticipation model letting the teacher recognition model instructing the student anticipation model on the most discriminative features, 2) It allows to leverage both labeled and unlabeled examples for training.
To study the contribution of each of these aspects, we assessed the performance of the proposed model when trained with different amounts of training data and in the presence or not of knowledge distillation.
In particular, we consider the following settings: 1) Supervised - training examples are formed sampling video clips exactly one second before the beginning of the action; 2) Augmented - training examples are obtained with temporal augmentation, i.e., we sample videos randomly and assign them the label of the future video segment located after one second. If no label appears in the future segment or if the label of the observed and future segments are the same, the example is discarded. Note that this sets contains all the supervised examples. 3) All - we consider all possible video clips, both labeled and unlabeled. Note that these settings can be considered only when knowledge distillation techniques are employed.

Table~\ref{tab:ablation_data} reports the results. It can be noted that just adding knowledge distillation techniques while keeping the amount of training data fixed always allows to improve performance. This suggests that knowledge distillation has a guiding effect which improves generalization. For instance, training the model with knowledge distillation relying only on supervised data allows to obtain a $+2.45$ in action performance when compared to training without knowledge distillation ($10.55$ vs $8.10$). 
Simply augmenting the number of training examples has a similar regularizing effect. Indeed, training the model with the augmented data allows to obtain an action score of $10.66$, which is larger by $+2.46$ with respect to the $8.10$ obtained considering only supervised data.
Combining both approaches allows to boost performance by $+4.73$ with respect to the base model ($12.83$ using all data and knowledge distillation vs $8.10$ using only supervised data and no knowledge distillation).
These results highlight how knowledge distillation has the effect of making training more data-efficient, allowing to increase the number of training examples and better exploit the available data.

\paragraph{Knowledge Distillation Loss}
To study the effectiveness of the proposed knowledge distillation loss, we compare its performance with respect to other known knowledge distillation losses. In particular, we compare with respect to the Knowledge Distillation (KD) loss proposed in~\cite{hinton2015distilling}, the Variational Information Distillation (VID) loss introduced in~\cite{ahn2019variational}, the Maximum Mean Discrepancy (MMD) loss~\cite{Huang2017}, and the ``Back To The Future'' (BTTF) loss described in in~\cite{tran2019back}. As baseline losses, we also consider Mean Squared Error (MSE), which assumes past and future representations to be spatio-temporally alined, and Global Average Pooling (GAP) + MSE, which applies the MSE loss to average pooling past and future losses.
As can be seen from Table~\ref{tab:ablation_loss},  all distillation approaches allow to improve results over the baseline model which does not consider knowledge distillation (first row). 
Among the compared approaches, the proposed loss achieves best action performance ($12.83$).
Combining GAP with MSE, allows to improve average performance ($19.99$ vs $19.71$), with a particular advantage over verbs ($23.94$ vs $22.72$) and a decrease in performance over nouns. 
In average, the proposed approach outperforms all competitors.

\begin{table}[t]
\centering
\adjustbox{max width=0.8\linewidth}{
\setlength{\tabcolsep}{2.5pt}
\renewcommand{\arraystretch}{1.2}
\begin{tabular}{lrrrr}
  \hline
\textbf{Method} & \textbf{VERB} & \textbf{NOUN} & \textbf{ACT.} & \textbf{AVG.} \\ 
  \hline
  No Distillation & 15.31 & 17.09 & 8.10 & 13.50 \\ 
  \hline
  KD\cite{hinton2015distilling} & 21.10 & 23.41 & 11.49 & 18.67 \\ 
  VID\cite{ahn2019variational} & 21.50 & 23.89 & 11.71 & 19.03 \\ 
  MMD\cite{Huang2017} & 21.56 & 22.96 & 11.93 & 18.82 \\ 
  BTTF\cite{tran2019back} & 21.51 & 24.02 & 12.04 & 19.19 \\ 
  \hline
  MSE & 22.72 & \textbf{24.67} & 11.73 & 19.71 \\ 
  GAP + MSE & \textbf{23.94} & 23.62 & \underline{12.42} & \underline{19.99} \\ 
  \hline
  Proposed & \underline{23.67} & \underline{24.43} & \textbf{12.83} & \textbf{20.31} \\ 
  \hline
\end{tabular}}
\caption{Comparison of Knowledge Distillation Loss Functions}
\label{tab:ablation_loss}
\end{table}

\subsection{Qualitative Examples}
\begin{figure}
    \centering
    \includegraphics[width=\linewidth]{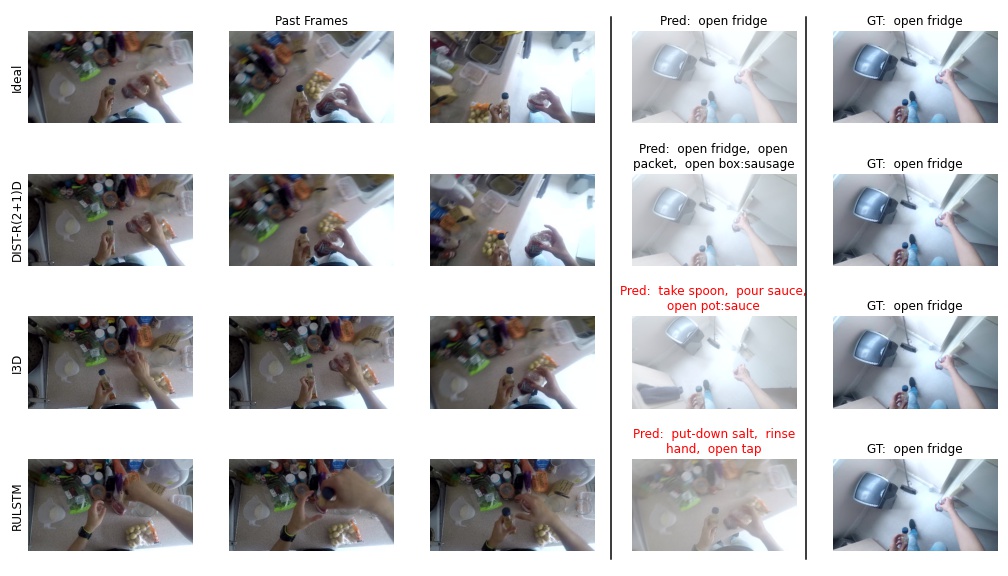}
    \caption{Different methods make their prediction based on different visual inputs due to their different runtime. See text for discussion and Appendix~\ref{appendix:qualitative} for additional qualitative examples.}
    \label{fig:qualitative}
\end{figure}

Figure~\ref{fig:qualitative} reports a qualitative example comparing the predictions of the proposed DIST-R(2+1)D, I3D and RULSTM on EPIC-KITCHENS-55. 
Due to their different runtimes, the methods observe different input clips to predict future actions at a given timestamp. 
The first three columns of Figure~\ref{fig:qualitative} report three frames from the video segment observed by the models. The fourth column reports the video frame appearing one second after the end of the observed segment. This frame is reported for reference and not observed by the model. The last column reports the (unobserved) first frame of the action to be predicted. 
The first row reports frames from the input clip which a method would ideally observe if it had zero runtime. 
The second, third and fourth rows are related to the three compared methods.
We report the Ground Truth action (GT) above the images in the last column and Top-3 predictions above the images in the fourth column. Wrong predictions are reported in red.

In the shown example, only the proposed DIST-R(2+1)D method correctly anticipates the ``open fridge'' action. This is likely due to the input observation which is more aligned with the ideal one (first row) than the other methods. 
Note that RULSTM predicts reasonable future actions for the observed segment (e.g., ``put-down salt'' as salt is manipulated), but it fails to anticipate the actual future action due to the misaligned video observation.
See Appendix~\ref{appendix:qualitative} for additional qualitative examples.

\section{Conclusion}
\label{sec:conclusion}
In this paper, we presented a novel evaluation protocol for egocentric action anticipation which explicitly considers a streaming scenario in which the timeliness of predictions is crucial.
We benchmarked different state-of-the-art anticipation approach and shown that classic offline evaluations are limited when algorithms need to be deployed on real hardware.
Noting that a small runtime is crucial for streaming anticipation, we proposed a method based on a lightweight 3D CNN and optimized with knowledge distillation techniques which achieved state-of-the-art performance on two datasets.
We believe that the proposed investigation can be useful in practical scenarios in which the evaluated models need to be deployed on real hardware.

\section*{Acknowledgment}
This research has been supported by MIUR AIM - Attrazione e Mobilit\`a Internazionale Linea 1 - AIM1893589 - CUP: E64118002540007, by MISE - PON I\&C 2014-2020 - Progetto ENIGMA - Prog n. F/190050/02/X44 – CUP: B61B19000520008, and by the project MEGABIT - PIAno di inCEntivi per la RIcerca di Ateneo 2020/2022 (PIACERI) – linea di intervento 2, DMI - University of Catania.

\appendix
\section{Implementation Details}
\label{appendix:implementation}
In this section, we report the implementation details of the proposed and compared methods. We trained all models on a server equipped with $4$ NVIDIA V100 GPUs. 

\subsection{Proposed Method}
We set $\lambda_d=20$ and $\lambda_c=1$ of the distillation loss of Eq. (2) of the main paper in all our experiments.
We follow the procedure described in the main paper and train our model on both labeled and unlabeled data.
Specifically, we sample video pairs at all possible timestamps $t$ within all training videos, which accounts to about $3.5M$, $8M$, and $2.3M$ training examples on EPIC-KITCHENS-55 and EGTEA Gaze+ respectively.
To maximize the amount of labeled data, we consider a training example to be labeled if at least half of the frames of the future observation are associated to an action segment.
When more labels are included in a future observation (action segments may overlap), we associate it with the most frequent one.
If a past and a future observation contain the same action, we consider the example as unlabeled as we may be sampling in the middle of a long action.
Since training sets obtained in these settings are very large and partly redundant, we found models to converge in one epoch.
At the beginning of a processed video, the computed $t_i^*$ value may be negative. In this case it is not possible to obtain an anticipated prediction for action $A_i$ and hence we predict $\hat y_{t_i^*}$ with a random guess.
We trained all models using the Adam optimizer~\cite{kingma2014adam} with a base learning rate of $1e-4$ and batch sizes equal to $80$.
The teacher models are fine-tuned from Kinetics pre-trained weights provided in the PyTorch library.
During training of both teacher and student models, we perform random horizontal flip and resize the input video so that the shortest side is equal to $64$. After resizing the clip, we perform a $64 \times 64$ random crop.
All these spatial augmentations are performed coherently on both past and future video clips included in the training example.
During test, we remove the random horizontal clip and replace the random crop with a center crop.
After training, we obtain our final model by averaging the weights of the $5$ best-performing checkpoints.

\subsection{Compared Methods} 
\label{sec:details_compared}
We train RULSTM using the official code and pre-computed features provided by the authors\footnote{\url{https://github.com/fpv-iplab/rulstm}}~\cite{furnari2020rolling}. For ED, we use RGB and optical flow features provided in~\cite{furnari2020rolling}. The TSN model has been trained using the Verb-Noun Marginal Cross Entropy loss proposed in~\cite{furnari2018leveraging} and adopting the suggested hyper-parameters. The LSTM baseline is trained using the same codebase of~\cite{furnari2020rolling} and same hyperparameters. The X3D-XS and SlowFast models are trained using the official code provided by the authors~\cite{feichtenhofer2019slowfast} and adopting the suggested parameters, finetuning from Kinetics-pretrained weights on 3 NVIDIA V100 GPUs. For I3D, we use a publicly available PyTorch implementation\footnote{https://github.com/piergiaj/pytorch-i3d}. The model is trained for $60$ epochs, with stochastic gradient descent and a base learning rate of $0.1$, which is multiplied by $0.1$ every $20$ epochs. The model is then finetuned from Kinetics-pretrained weights using 3 NVIDIA V100 GPUs and a batch size of $24$. We train the R(2+1)D-based models following the same parameters as the proposed approach. 

\section{Qualitative Examples}
\label{appendix:qualitative}
\begin{figure*}
    \centering
    \includegraphics[width=\linewidth]{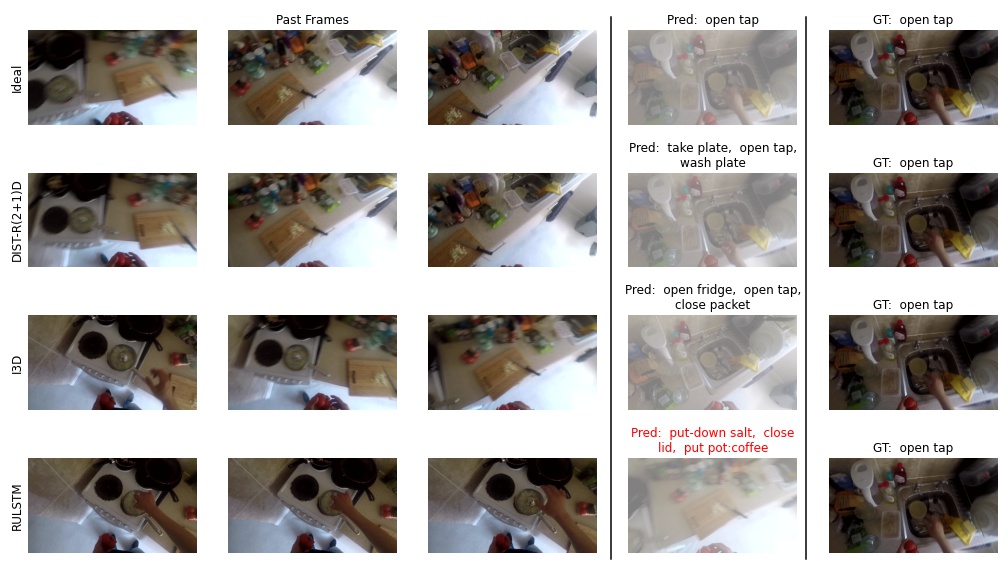}
    \caption{Qualitative examples.}
    \label{fig:qualitative_1}
\end{figure*}

\begin{figure*}
    \centering
    \includegraphics[width=\linewidth]{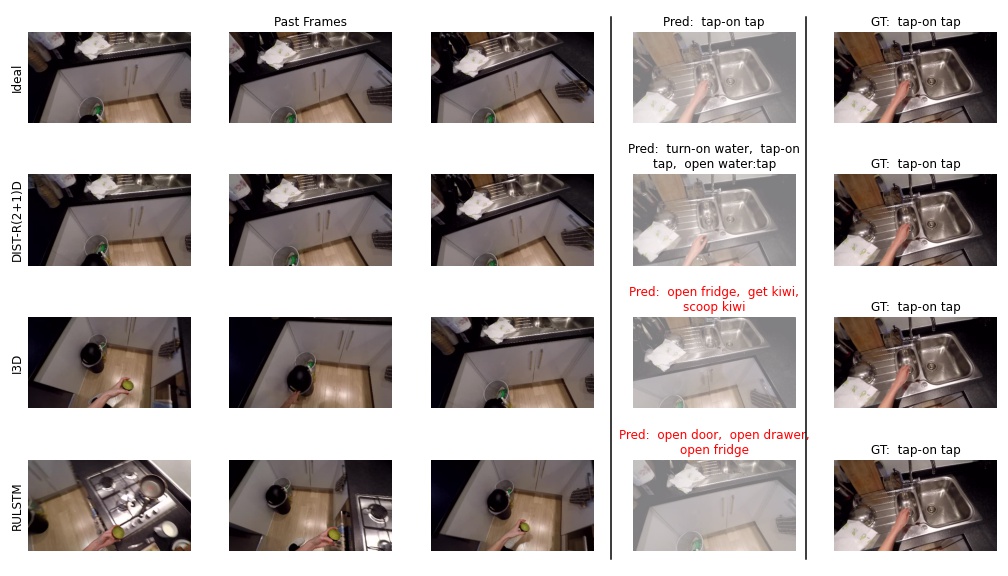}
    \caption{Qualitative examples.}
    \label{fig:qualitative_2}
\end{figure*}

\begin{figure*}
    \centering
    \includegraphics[width=\linewidth]{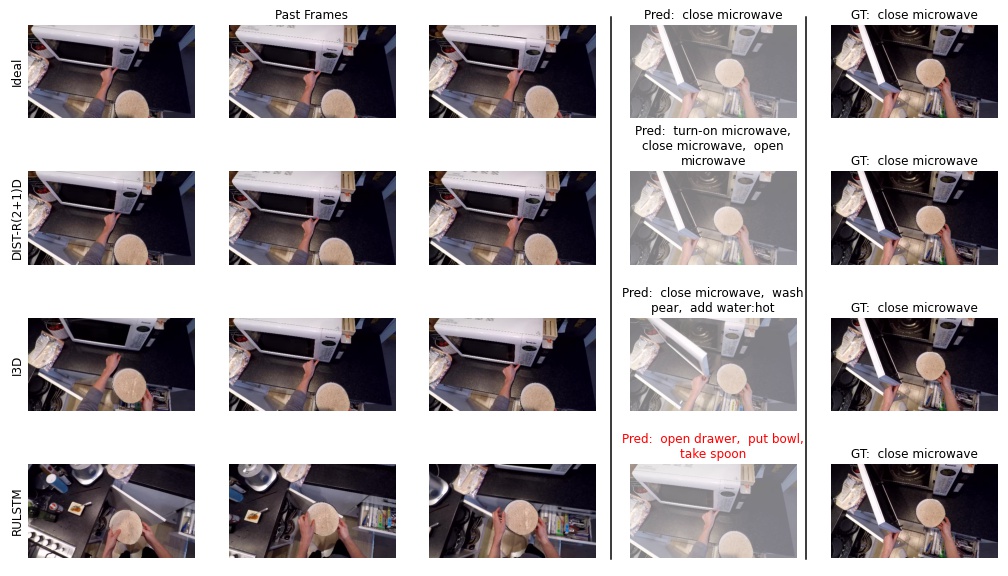}
    \caption{Qualitative examples.}
    \label{fig:qualitative_3}
\end{figure*}

\begin{figure*}
    \centering
    \includegraphics[width=\linewidth]{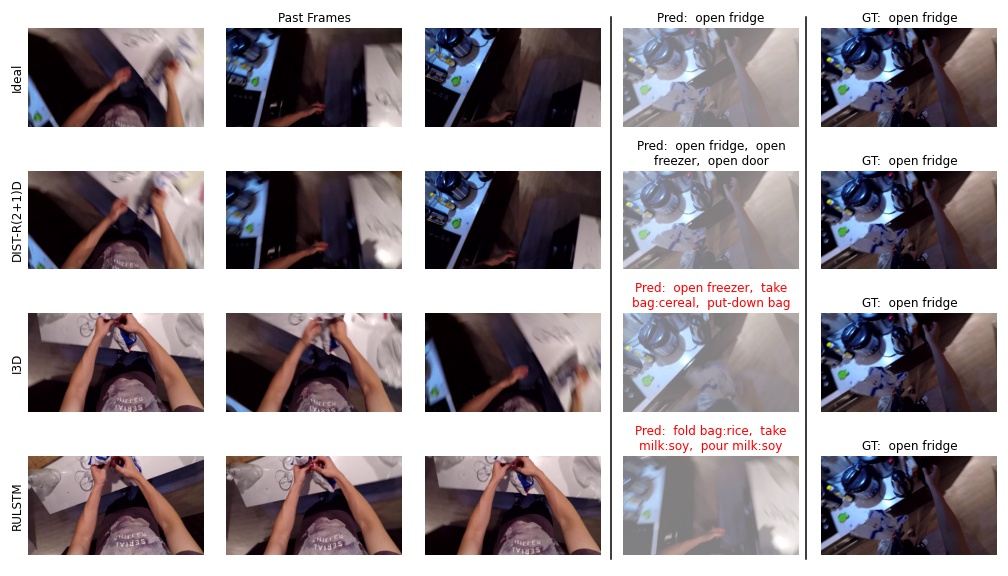}
    \caption{Qualitative examples.}
    \label{fig:qualitative_4}
\end{figure*}

\begin{figure*}
    \centering
    \includegraphics[width=\linewidth]{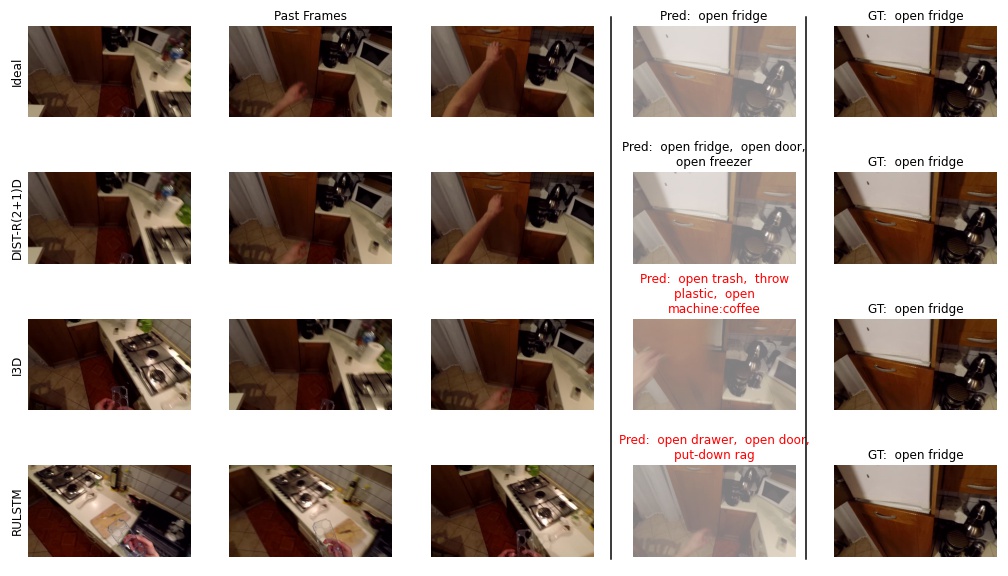}
    \caption{Qualitative examples.}
    \label{fig:qualitative_5}
\end{figure*}

\begin{figure*}
    \centering
    \includegraphics[width=\linewidth]{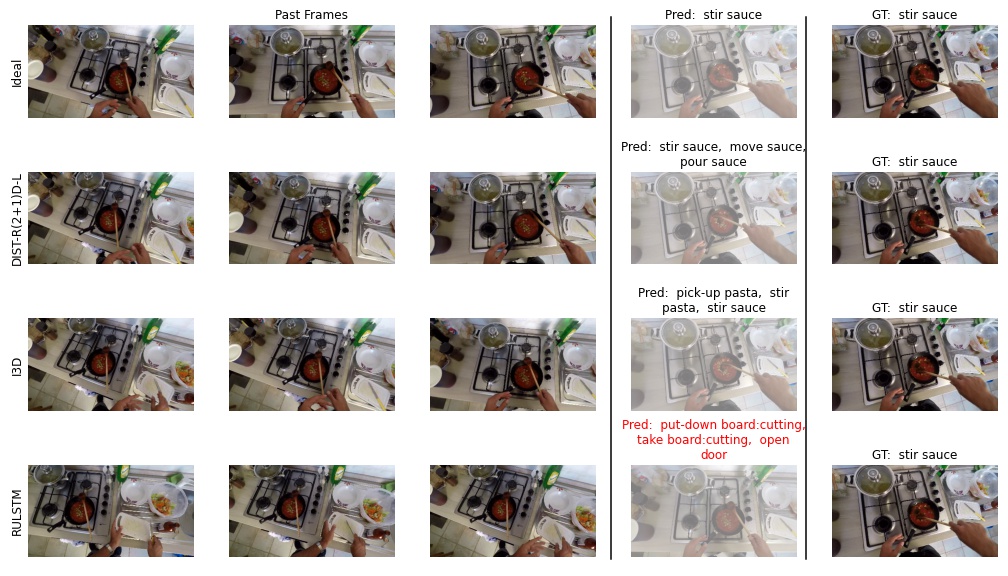}
    \caption{Qualitative examples.}
    \label{fig:qualitative_6}
\end{figure*}

\begin{figure*}
    \centering
    \includegraphics[width=\linewidth]{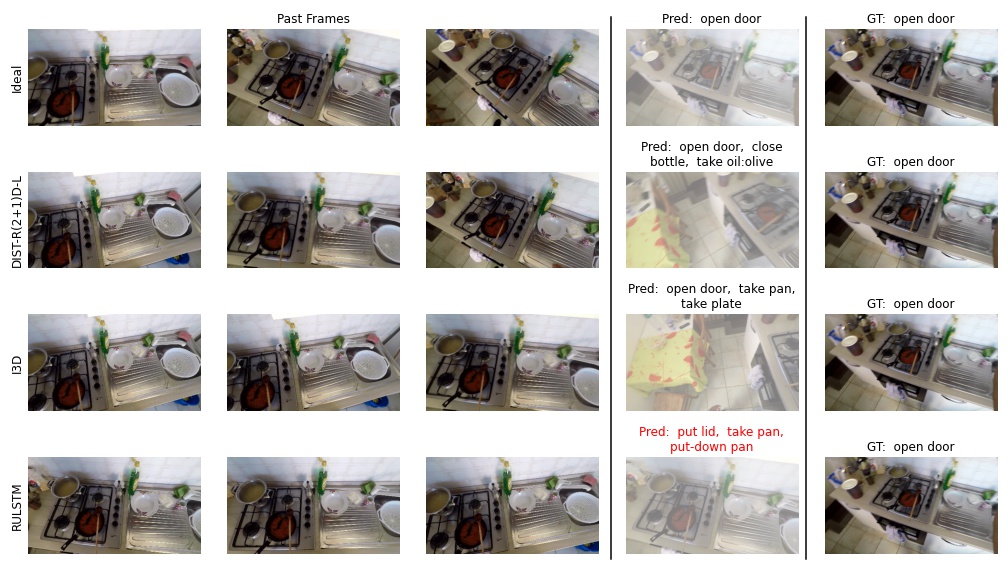}
    \caption{Qualitative examples.}
    \label{fig:qualitative_7}
\end{figure*}

\begin{figure*}
    \centering
    \includegraphics[width=\linewidth]{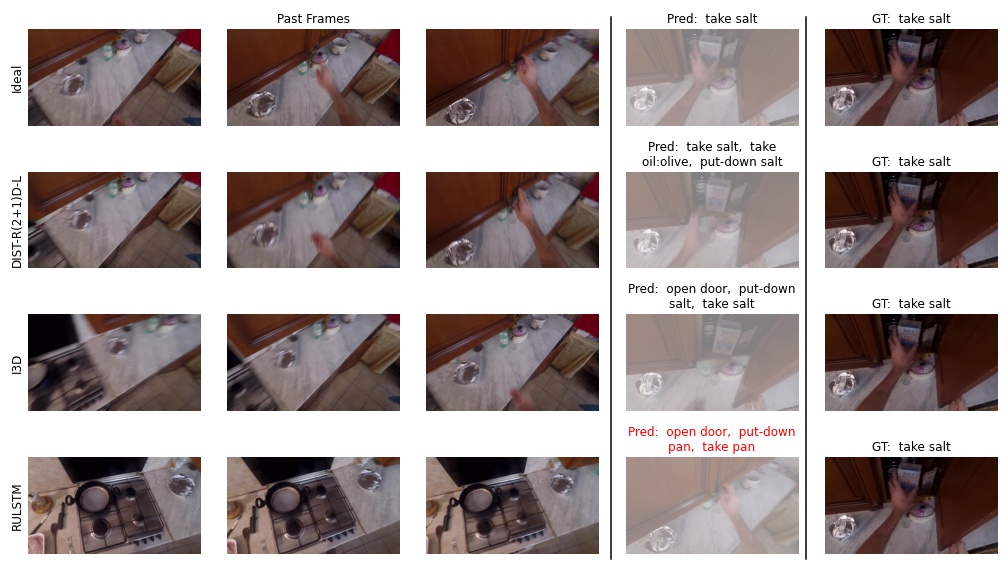}
    \caption{Qualitative examples.}
    \label{fig:qualitative_8}
\end{figure*}

\begin{figure*}
    \centering
    \includegraphics[width=\linewidth]{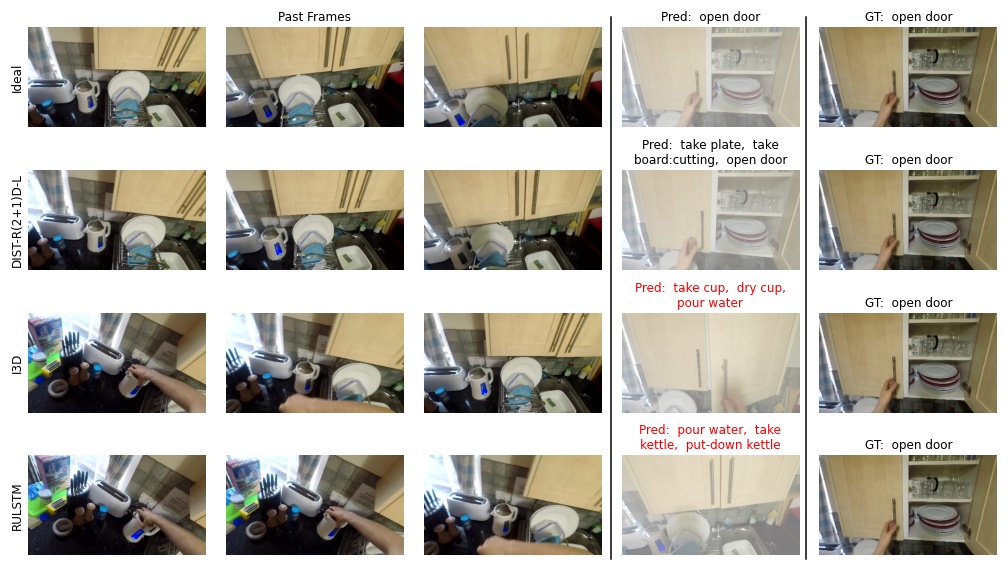}
    \caption{Qualitative examples.}
    \label{fig:qualitative_9}
\end{figure*}

\begin{figure*}
    \centering
    \includegraphics[width=\linewidth]{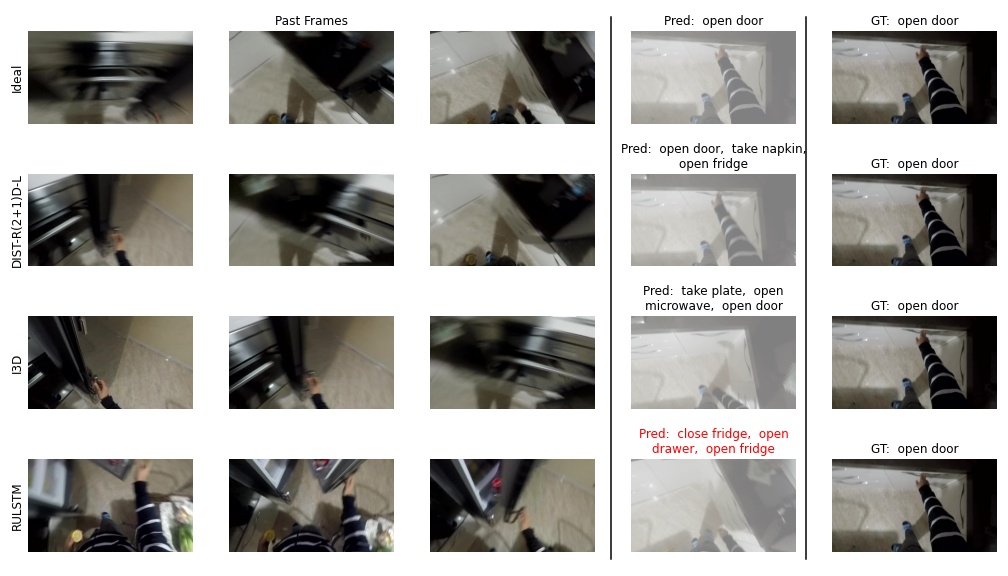}
    \caption{Qualitative examples.}
    \label{fig:qualitative_10}
\end{figure*}

\begin{figure*}
    \centering
    \includegraphics[width=\linewidth]{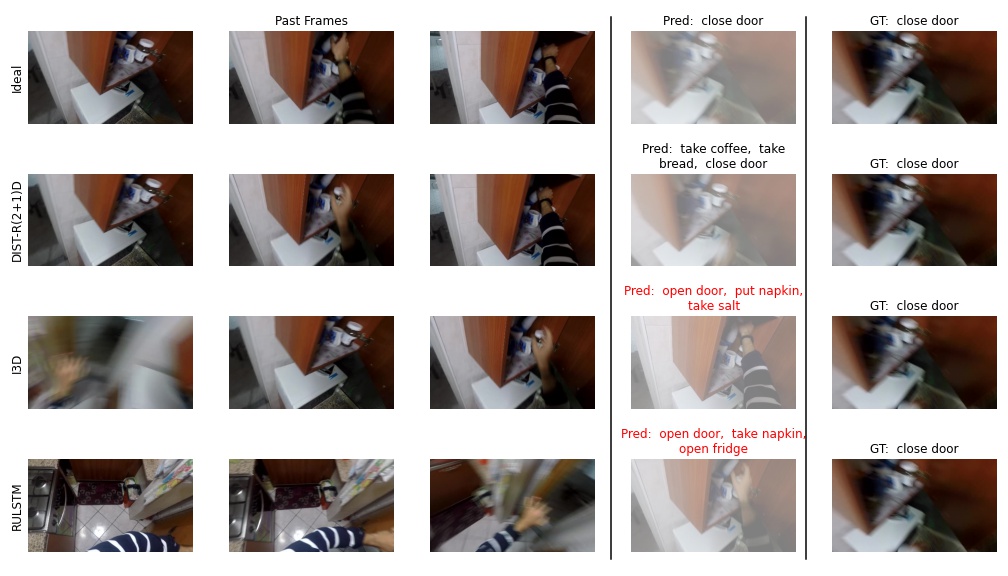}
    \caption{Qualitative examples.}
    \label{fig:qualitative_11}
\end{figure*}

Figures~\ref{fig:qualitative_1}-\ref{fig:qualitative_11} report additional qualitative examples. 
As can be noted, RULSTM usually observes a significantly different input sequence, which makes predictions less effective 

(e.g., Figures~\ref{fig:qualitative_1},~\ref{fig:qualitative_3},~\ref{fig:qualitative_8},~\ref{fig:qualitative_10},~\ref{fig:qualitative_11}).
Despite DIST-R(2+1)D and I3D tend to observe similar inputs, the former tends to make better predictions than the latter (e.g., Figures~\ref{fig:qualitative_2},~\ref{fig:qualitative_9},~\ref{fig:qualitative_11}).

{\small
\bibliographystyle{ieee}
\bibliography{egbib}

\begin{thebibliography}{10}\itemsep=-1pt

\bibitem{ahn2019variational}
S.~Ahn, S.~X. Hu, A.~Damianou, N.~D. Lawrence, and Z.~Dai.
\newblock Variational information distillation for knowledge transfer.
\newblock In {\em Proceedings of the IEEE/CVF Conference on Computer Vision and
  Pattern Recognition}, pages 9163--9171, 2019.

\bibitem{camporese2020knowledge}
G.~Camporese, P.~Coscia, A.~Furnari, G.~M. Farinella, and L.~Ballan.
\newblock Knowledge distillation for action anticipation via label smoothing.
\newblock In {\em International Conference on Pattern Recognition (ICPR)},
  2020.

\bibitem{carreira2017quo}
J.~Carreira and A.~Zisserman.
\newblock Quo vadis, action recognition? a new model and the kinetics dataset.
\newblock In {\em 2017 IEEE Conference on Computer Vision and Pattern
  Recognition (CVPR)}, pages 4724--4733, 2017.

\bibitem{damen2020epic}
D.~Damen, H.~Doughty, G.~Farinella, S.~Fidler, A.~Furnari, E.~Kazakos,
  D.~Moltisanti, J.~Munro, T.~Perrett, W.~Price, and M.~Wray.
\newblock The epic-kitchens dataset: Collection, challenges and baselines.
\newblock {\em IEEE Transactions on Pattern Analysis and Machine Intelligence},
  pages 1--1, 2020.

\bibitem{de2016online}
R.~De~Geest, E.~Gavves, A.~Ghodrati, Z.~Li, C.~Snoek, and T.~Tuytelaars.
\newblock Online action detection.
\newblock In {\em European Conference on Computer Vision}, pages 269--284.
  Springer, 2016.

\bibitem{dessalene2021forecasting}
E.~Dessalene, C.~Devaraj, M.~Maynord, C.~Fermuller, and Y.~Aloimonos.
\newblock Forecasting action through contact representations from first person
  video.
\newblock {\em IEEE Transactions on Pattern Analysis and Machine Intelligence},
  2021.

\bibitem{feichtenhofer2020x3d}
C.~Feichtenhofer.
\newblock X3d: Expanding architectures for efficient video recognition.
\newblock In {\em Proceedings of the IEEE/CVF Conference on Computer Vision and
  Pattern Recognition}, pages 203--213, 2020.

\bibitem{feichtenhofer2019slowfast}
C.~Feichtenhofer, H.~Fan, J.~Malik, and K.~He.
\newblock Slowfast networks for video recognition.
\newblock In {\em Proceedings of the IEEE/CVF International Conference on
  Computer Vision}, pages 6202--6211, 2019.

\bibitem{Fernando21}
B.~Fernando and S.~Herath.
\newblock {Anticipating human actions by correlating past with the future with
  Jaccard similarity measures}.
\newblock {\em CVPR}, 2021.

\bibitem{furnari2017next}
A.~Furnari, S.~Battiato, K.~Grauman, and G.~M. Farinella.
\newblock Next-active-object prediction from egocentric videos.
\newblock {\em Journal of Visual Communication and Image Representation},
  49:401--411, 2017.

\bibitem{furnari2018leveraging}
A.~Furnari, S.~Battiato, and G.~Maria~Farinella.
\newblock Leveraging uncertainty to rethink loss functions and evaluation
  measures for egocentric action anticipation.
\newblock In {\em Proceedings of the European Conference on Computer Vision
  (ECCV) Workshops}, pages 0--0, 2018.

\bibitem{furnari2020rolling}
A.~Furnari and G.~Farinella.
\newblock Rolling-unrolling lstms for action anticipation from first-person
  video.
\newblock {\em IEEE transactions on pattern analysis and machine intelligence},
  2020.

\bibitem{gao2017red}
J.~Gao, Z.~Yang, and R.~Nevatia.
\newblock Red: Reinforced encoder-decoder networks for action anticipation.
\newblock In {\em British Machine Vision Conference}, 2017.

\bibitem{gers1999learning}
F.~Gers, J.~Schmidhuber, and F.~Cummins.
\newblock Learning to forget: continual prediction with lstm.
\newblock In {\em 1999 Ninth International Conference on Artificial Neural
  Networks ICANN 99. (Conf. Publ. No. 470)}, volume~2, pages 850--855 vol.2,
  1999.

\bibitem{he2015spatial}
K.~He, X.~Zhang, S.~Ren, and J.~Sun.
\newblock Spatial pyramid pooling in deep convolutional networks for visual
  recognition.
\newblock {\em IEEE transactions on pattern analysis and machine intelligence},
  37(9):1904--1916, 2015.

\bibitem{he2016deep}
K.~He, X.~Zhang, S.~Ren, and J.~Sun.
\newblock Deep residual learning for image recognition.
\newblock In {\em Proceedings of the IEEE conference on computer vision and
  pattern recognition}, pages 770--778, 2016.

\bibitem{hinton2015distilling}
G.~Hinton, O.~Vinyals, and J.~Dean.
\newblock Distilling the knowledge in a neural network.
\newblock {\em arXiv preprint arXiv:1503.02531}, 2015.

\bibitem{hoai2014max}
M.~Hoai and F.~De~la Torre.
\newblock Max-margin early event detectors.
\newblock {\em International Journal of Computer Vision}, 107(2):191--202,
  2014.

\bibitem{howard2017mobilenets}
A.~G. Howard, M.~Zhu, B.~Chen, D.~Kalenichenko, W.~Wang, T.~Weyand,
  M.~Andreetto, and H.~Adam.
\newblock Mobilenets: Efficient convolutional neural networks for mobile vision
  applications.
\newblock {\em arXiv preprint arXiv:1704.04861}, 2017.

\bibitem{Huang2017}
Z.~Huang and N.~Wang.
\newblock {Like what you like: Knowledge distill via neuron selectivity
  transfer}, 2017.

\bibitem{iandola2016squeezenet}
F.~N. Iandola, S.~Han, M.~W. Moskewicz, K.~Ashraf, W.~J. Dally, and K.~Keutzer.
\newblock Squeezenet: Alexnet-level accuracy with 50x fewer parameters and
  \textless 0.5 mb model size.
\newblock {\em arXiv preprint arXiv:1602.07360}, 2016.

\bibitem{kanade2012first}
T.~Kanade and M.~Hebert.
\newblock First-person vision.
\newblock {\em Proceedings of the IEEE}, 100(8):2442--2453, 2012.

\bibitem{kingma2014adam}
D.~P. Kingma and J.~Ba.
\newblock Adam: A method for stochastic optimization.
\newblock {\em arXiv preprint arXiv:1412.6980}, 2014.

\bibitem{koppula2015anticipating}
H.~S. Koppula and A.~Saxena.
\newblock Anticipating human activities using object affordances for reactive
  robotic response.
\newblock {\em IEEE transactions on pattern analysis and machine intelligence},
  38(1):14--29, 2015.

\bibitem{kristan2017visual}
M.~Kristan, A.~Leonardis, J.~Matas, M.~Felsberg, R.~Pflugfelder,
  L.~ˇCehovin~Zajc, T.~Vojir, G.~Hager, A.~Lukezic, A.~Eldesokey, et~al.
\newblock The visual object tracking vot2017 challenge results.
\newblock In {\em Proceedings of the IEEE international conference on computer
  vision workshops}, pages 1949--1972, 2017.

\bibitem{li2020towards}
M.~Li, Y.-X. Wang, and D.~Ramanan.
\newblock Towards streaming perception.
\newblock In {\em European Conference on Computer Vision}, pages 473--488.
  Springer, 2020.

\bibitem{li2021eye}
Y.~Li, M.~Liu, and J.~Rehg.
\newblock In the eye of the beholder: Gaze and actions in first person video.
\newblock {\em IEEE Transactions on Pattern Analysis and Machine Intelligence},
  2021.

\bibitem{lin2017focal}
T.-Y. Lin, P.~Goyal, R.~Girshick, K.~He, and P.~Doll{\'a}r.
\newblock Focal loss for dense object detection.
\newblock In {\em Proceedings of the IEEE international conference on computer
  vision}, pages 2980--2988, 2017.

\bibitem{liu2020forecasting}
M.~Liu, S.~Tang, Y.~Li, and J.~M. Rehg.
\newblock Forecasting human-object interaction: joint prediction of motor
  attention and actions in first person video.
\newblock In {\em European Conference on Computer Vision}, pages 704--721.
  Springer, 2020.

\bibitem{liu2016ssd}
W.~Liu, D.~Anguelov, D.~Erhan, C.~Szegedy, S.~Reed, C.-Y. Fu, and A.~C. Berg.
\newblock Ssd: Single shot multibox detector.
\newblock In {\em European conference on computer vision}, pages 21--37.
  Springer, 2016.

\bibitem{miech2019leveraging}
A.~Miech, I.~Laptev, J.~Sivic, H.~Wang, L.~Torresani, and D.~Tran.
\newblock Leveraging the present to anticipate the future in videos.
\newblock In {\em Proceedings of the IEEE/CVF Conference on Computer Vision and
  Pattern Recognition Workshops}, pages 0--0, 2019.

\bibitem{passalis2020heterogeneous}
N.~Passalis, M.~Tzelepi, and A.~Tefas.
\newblock Heterogeneous knowledge distillation using information flow modeling.
\newblock In {\em Proceedings of the IEEE/CVF Conference on Computer Vision and
  Pattern Recognition}, pages 2339--2348, 2020.

\bibitem{pytorch}
A.~Paszke, S.~Gross, F.~Massa, A.~Lerer, J.~Bradbury, G.~Chanan, T.~Killeen,
  Z.~Lin, N.~Gimelshein, L.~Antiga, A.~Desmaison, A.~Kopf, E.~Yang, Z.~DeVito,
  M.~Raison, A.~Tejani, S.~Chilamkurthy, B.~Steiner, L.~Fang, J.~Bai, and
  S.~Chintala.
\newblock Pytorch: An imperative style, high-performance deep learning library.
\newblock In H.~Wallach, H.~Larochelle, A.~Beygelzimer, F.~d\textquotesingle
  Alch\'{e}-Buc, E.~Fox, and R.~Garnett, editors, {\em Advances in Neural
  Information Processing Systems 32}, pages 8024--8035. Curran Associates,
  Inc., 2019.

\bibitem{qi2021self}
Z.~Qi, S.~Wang, C.~Su, L.~Su, Q.~Huang, and Q.~Tian.
\newblock Self-regulated learning for egocentric video activity anticipation.
\newblock {\em IEEE Transactions on Pattern Analysis and Machine Intelligence},
  2021.

\bibitem{rastegari2016xnor}
M.~Rastegari, V.~Ordonez, J.~Redmon, and A.~Farhadi.
\newblock Xnor-net: Imagenet classification using binary convolutional neural
  networks.
\newblock In {\em European conference on computer vision}, pages 525--542.
  Springer, 2016.

\bibitem{redmon2016you}
J.~Redmon, S.~Divvala, R.~Girshick, and A.~Farhadi.
\newblock You only look once: Unified, real-time object detection.
\newblock In {\em Proceedings of the IEEE conference on computer vision and
  pattern recognition}, pages 779--788, 2016.

\bibitem{redmon2017yolo9000}
J.~Redmon and A.~Farhadi.
\newblock Yolo9000: better, faster, stronger.
\newblock In {\em Proceedings of the IEEE conference on computer vision and
  pattern recognition}, pages 7263--7271, 2017.

\bibitem{ren2016faster}
S.~Ren, K.~He, R.~Girshick, and J.~Sun.
\newblock Faster r-cnn: towards real-time object detection with region proposal
  networks.
\newblock {\em IEEE transactions on pattern analysis and machine intelligence},
  39(6):1137--1149, 2016.

\bibitem{romero2014fitnets}
A.~Romero, N.~Ballas, S.~E. Kahou, A.~Chassang, C.~Gatta, and Y.~Bengio.
\newblock Fitnets: Hints for thin deep nets.
\newblock {\em arXiv preprint arXiv:1412.6550}, 2014.

\bibitem{sadegh2017encouraging}
M.~Sadegh~Aliakbarian, F.~Sadat~Saleh, M.~Salzmann, B.~Fernando, L.~Petersson,
  and L.~Andersson.
\newblock Encouraging lstms to anticipate actions very early.
\newblock In {\em Proceedings of the IEEE International Conference on Computer
  Vision}, pages 280--289, 2017.

\bibitem{sener2020temporal}
F.~Sener, D.~Singhania, and A.~Yao.
\newblock Temporal aggregate representations for long-range video
  understanding.
\newblock In {\em European Conference on Computer Vision}, pages 154--171.
  Springer, 2020.

\bibitem{simonyan2014very}
K.~Simonyan and A.~Zisserman.
\newblock Very deep convolutional networks for large-scale image recognition.
\newblock In {\em International Conference on Learning Representations}, 2014.

\bibitem{soran2015generating}
B.~Soran, A.~Farhadi, and L.~Shapiro.
\newblock Generating notifications for missing actions: Don't forget to turn
  the lights off!
\newblock In {\em Proceedings of the IEEE International Conference on Computer
  Vision}, pages 4669--4677, 2015.

\bibitem{szegedy2015going}
C.~Szegedy, W.~Liu, Y.~Jia, P.~Sermanet, S.~Reed, D.~Anguelov, D.~Erhan,
  V.~Vanhoucke, and A.~Rabinovich.
\newblock Going deeper with convolutions.
\newblock In {\em Proceedings of the IEEE conference on computer vision and
  pattern recognition}, pages 1--9, 2015.

\bibitem{tan2019efficientnet}
M.~Tan and Q.~Le.
\newblock Efficientnet: Rethinking model scaling for convolutional neural
  networks.
\newblock In {\em International Conference on Machine Learning}, pages
  6105--6114. PMLR, 2019.

\bibitem{tran2018closer}
D.~Tran, H.~Wang, L.~Torresani, J.~Ray, Y.~LeCun, and M.~Paluri.
\newblock A closer look at spatiotemporal convolutions for action recognition.
\newblock In {\em Proceedings of the IEEE conference on Computer Vision and
  Pattern Recognition}, pages 6450--6459, 2018.

\bibitem{tran2019back}
V.~Tran, Y.~Wang, and M.~Hoai.
\newblock Back to the future: Knowledge distillation for human action
  anticipation.
\newblock {\em arXiv preprint arXiv:1904.04868}, 2019.

\bibitem{vondrick2016anticipating}
C.~Vondrick, H.~Pirsiavash, and A.~Torralba.
\newblock Anticipating visual representations from unlabeled video.
\newblock In {\em IEEE Conference on Computer Vision and Pattern Recognition},
  pages 98--106, 2016.

\bibitem{wang2016temporal}
L.~Wang, Y.~Xiong, Z.~Wang, Y.~Qiao, D.~Lin, X.~Tang, and L.~Van~Gool.
\newblock Temporal segment networks: Towards good practices for deep action
  recognition.
\newblock In {\em European conference on computer vision}, pages 20--36.
  Springer, 2016.

\bibitem{Wang2019}
X.~Wang, J.~F. Hu, J.~H. Lai, J.~Zhang, and W.~S. Zheng.
\newblock {Progressive teacher-student learning for early action prediction}.
\newblock In {\em Proceedings of the IEEE Computer Society Conference on
  Computer Vision and Pattern Recognition}, 2019.

\bibitem{wu2020learning}
Y.~Wu, L.~Zhu, X.~Wang, Y.~Yang, and F.~Wu.
\newblock Learning to anticipate egocentric actions by imagination.
\newblock {\em IEEE Transactions on Image Processing}, 30:1143--1152, 2020.

\bibitem{yim2017gift}
J.~Yim, D.~Joo, J.~Bae, and J.~Kim.
\newblock A gift from knowledge distillation: Fast optimization, network
  minimization and transfer learning.
\newblock In {\em Proceedings of the IEEE Conference on Computer Vision and
  Pattern Recognition}, pages 4133--4141, 2017.

\bibitem{zhang2020egocentric}
T.~Zhang, W.~Min, Y.~Zhu, Y.~Rui, and S.~Jiang.
\newblock An egocentric action anticipation framework via fusing intuition and
  analysis.
\newblock In {\em Proceedings of the 28th ACM International Conference on
  Multimedia}, pages 402--410, 2020.

\end{thebibliography}
}

\end{document}


\title{Towards Streaming Egocentric Action Anticipation\\(Supplementary Material)}

\author{First Author\\
Institution1\\
Institution1 address\\
{\tt\small firstauthor@i1.org}
\and
Second Author\\
Institution2\\
First line of institution2 address\\
{\tt\small secondauthor@i2.org}
}

\maketitle

This supplementary material reports the implementation details about the proposed method and the compared approaches in Section~\ref{sec:details}, as well as additional qualitative examples in Section~\ref{sec:qualitative}.

\section{Implementation Details}
\label{sec:details}
In this section, we report the implementation details of the proposed and compared methods. We trained all models on a server equipped with $4$ NVIDIA V100 GPUs. 

\subsection{Proposed Method}
\label{sec:details_propsoed}
We set $\lambda_d=20$ and $\lambda_c=1$ of the distillation loss of Eq. (2) of the main paper in all our experiments.
We follow the procedure described in the main paper and train our model on both labeled and unlabeled data.
Specifically, we sample video pairs at all possible timestamps $t$ within all training videos, which accounts to about $3.5M$, $8M$, and $2.3M$ training examples on EPIC-KITCHENS-55 and EGTEA Gaze+ respectively.
To maximize the amount of labeled data, we consider a training example to be labeled if at least half of the frames of the future observation are associated to an action segment.
When more labels are included in a future observation (action segments may overlap), we associate it with the most frequent one.
If a past and a future observation contain the same action, we consider the example as unlabeled as we may be sampling in the middle of a long action.
Since training sets obtained in these settings are very large and partly redundant, we found models to converge in one epoch.
At the beginning of a processed video, the computed $t_i^*$ value may be negative. In this case it is not possible to obtain an anticipated prediction for action $A_i$ and hence we predict $\hat y_{t_i^*}$ with a random guess.
We trained all models using the Adam optimizer~\cite{kingma2014adam} with a base learning rate of $1e-4$ and batch sizes equal to $80$.
The teacher models are fine-tuned from Kinetics pre-trained weights provided in the PyTorch library.
During training of both teacher and student models, we perform random horizontal flip and resize the input video so that the shortest side is equal to $64$. After resizing the clip, we perform a $64 \times 64$ random crop.
All these spatial augmentations are performed coherently on both past and future video clips included in the training example.
During test, we remove the random horizontal clip and replace the random crop with a center crop.
After training, we obtain our final model by averaging the weights of the $5$ best-performing checkpoints.

\subsection{Compared Methods} 
\label{sec:details_compared}
We train RULSTM using the official code and pre-computed features provided by the authors\footnote{\url{https://github.com/fpv-iplab/rulstm}}~\cite{furnari2020rolling}. For ED, we 
use RGB and optical flow features provided in~\cite{furnari2020rolling}. The TSN model has been trained using the Verb-Noun Marginal Cross Entropy loss proposed in~\cite{furnari2018leveraging} and adopting the suggested hyper-parameters. The LSTM baseline is trained using the same codebase of~\cite{furnari2020rolling} and same hyperparameters. The X3D-XS and SlowFast models are trained using the official code provided by the authors~\cite{feichtenhofer2019slowfast} and adopting the suggested parameters, finetuning from Kinetics-pretrained weights on 3 NVIDIA V100 GPUs. For I3D, we use a publicly available PyTorch implementation\footnote{https://github.com/piergiaj/pytorch-i3d}. The model is trained for $60$ epochs, with stochastic gradient descent and a base learning rate of $0.1$, which is multiplied by $0.1$ every $20$ epochs. The model is then finetuned from Kinetics-pretrained weights using 3 NVIDIA V100 GPUs and a batch size of $24$. We train the R(2+1)D-based models following the same parameters as the proposed approach. 

\section{Qualitative Examples}
\label{sec:qualitative}
\begin{figure*}
    \centering
    \includegraphics[width=\linewidth]{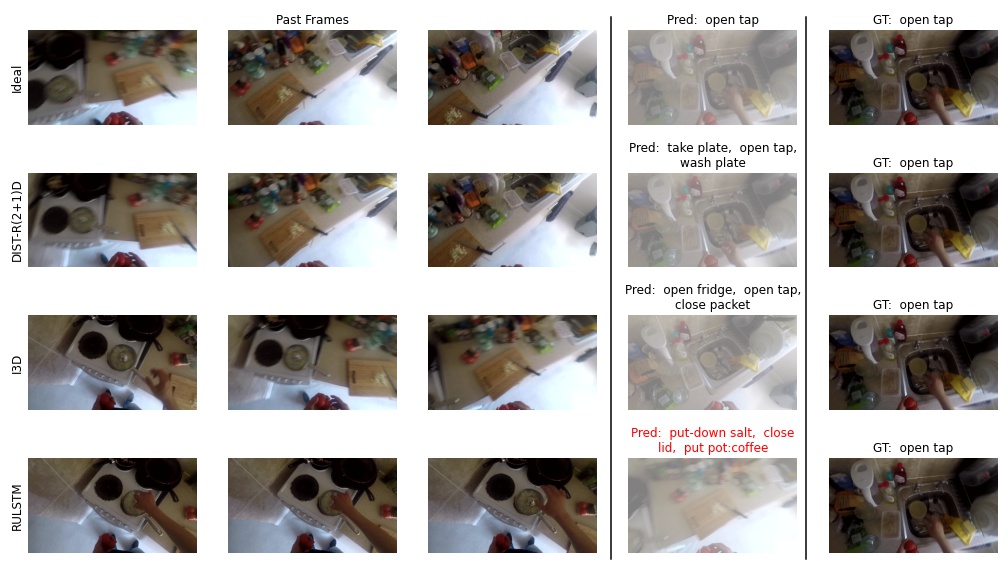}
    \caption{Qualitative examples.}
    \label{fig:qualitative_1}
\end{figure*}

\begin{figure*}
    \centering
    \includegraphics[width=\linewidth]{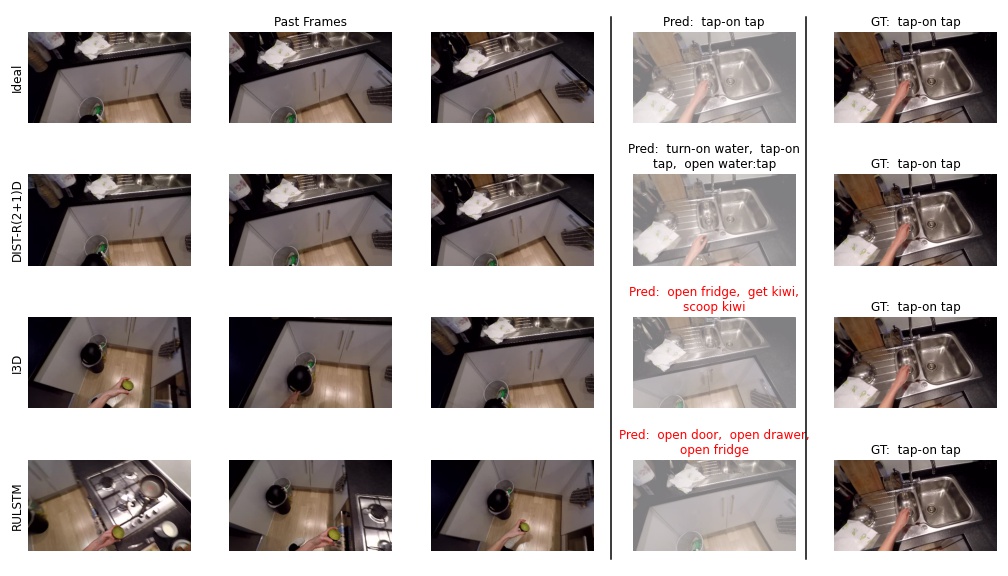}
    \caption{Qualitative examples.}
    \label{fig:qualitative_2}
\end{figure*}

\begin{figure*}
    \centering
    \includegraphics[width=\linewidth]{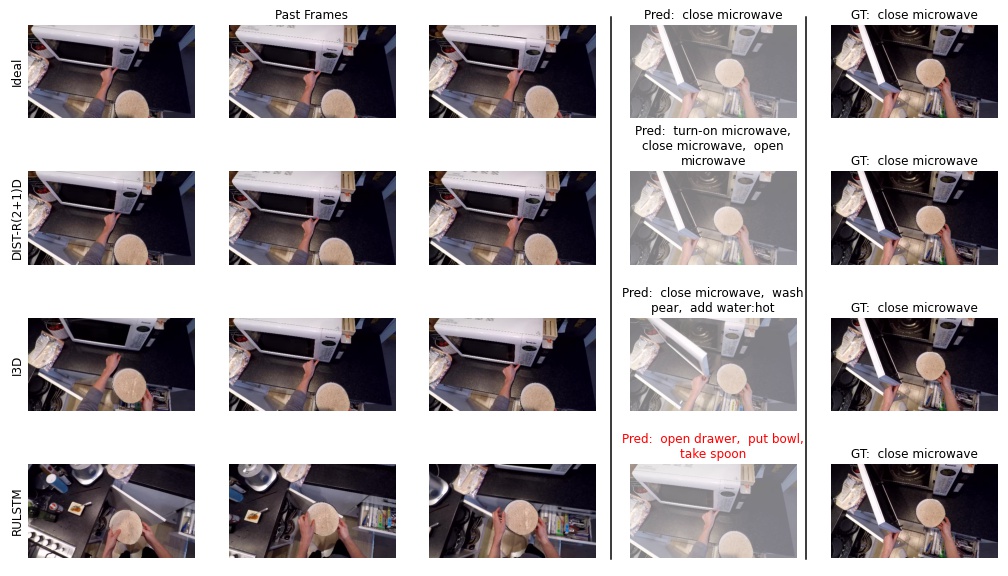}
    \caption{Qualitative examples.}
    \label{fig:qualitative_3}
\end{figure*}

\begin{figure*}
    \centering
    \includegraphics[width=\linewidth]{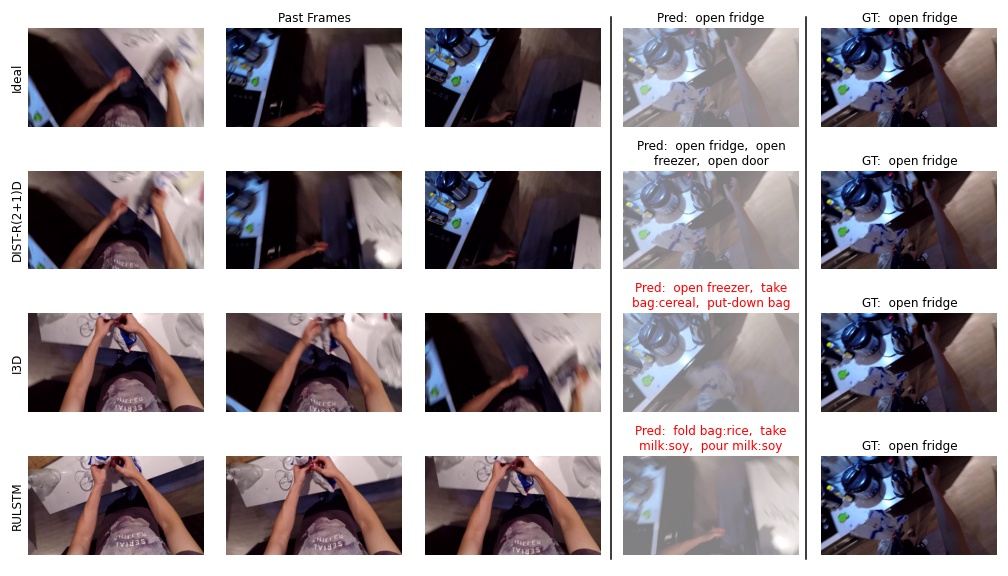}
    \caption{Qualitative examples.}
    \label{fig:qualitative_4}
\end{figure*}

\begin{figure*}
    \centering
    \includegraphics[width=\linewidth]{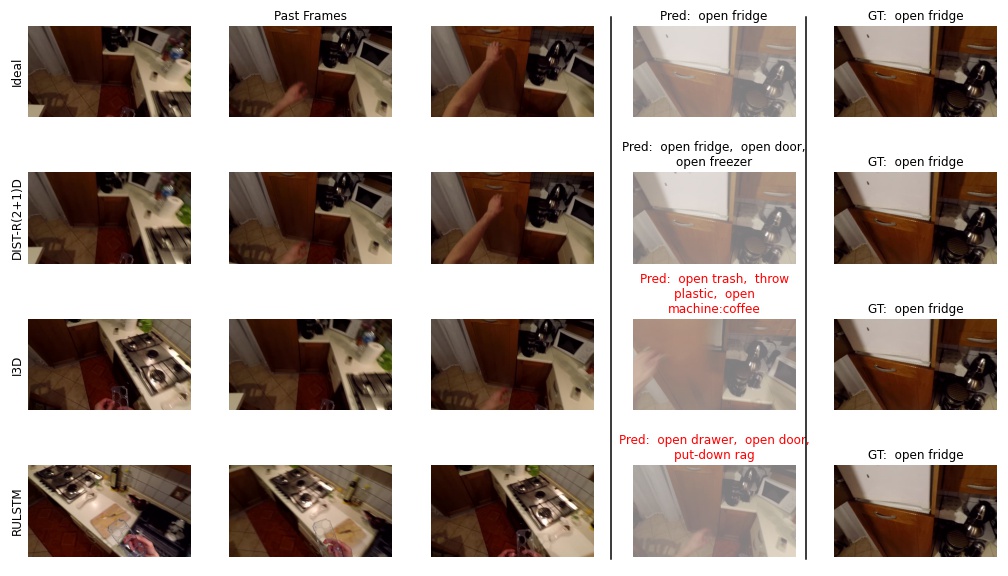}
    \caption{Qualitative examples.}
    \label{fig:qualitative_5}
\end{figure*}

\begin{figure*}
    \centering
    \includegraphics[width=\linewidth]{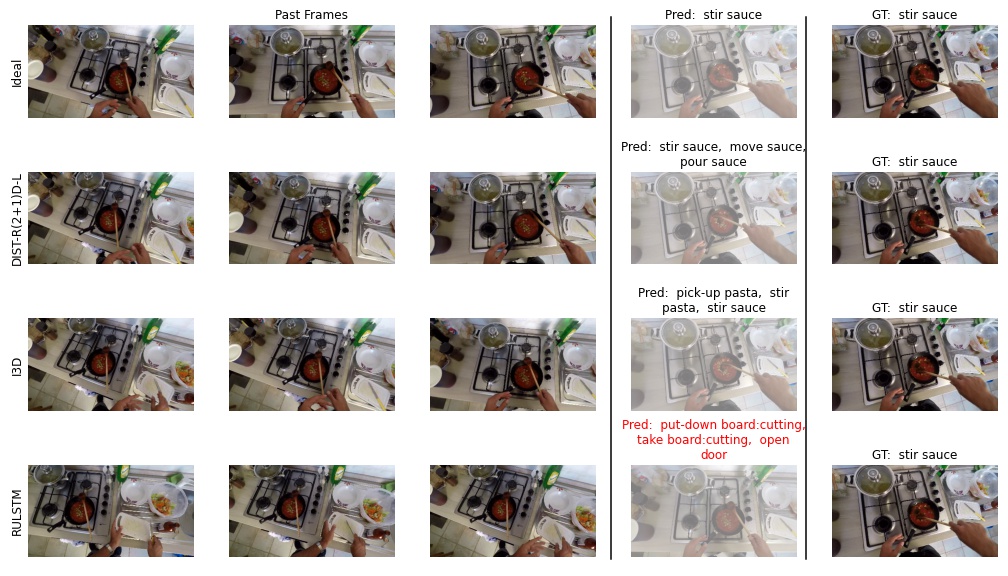}
    \caption{Qualitative examples.}
    \label{fig:qualitative_6}
\end{figure*}

\begin{figure*}
    \centering
    \includegraphics[width=\linewidth]{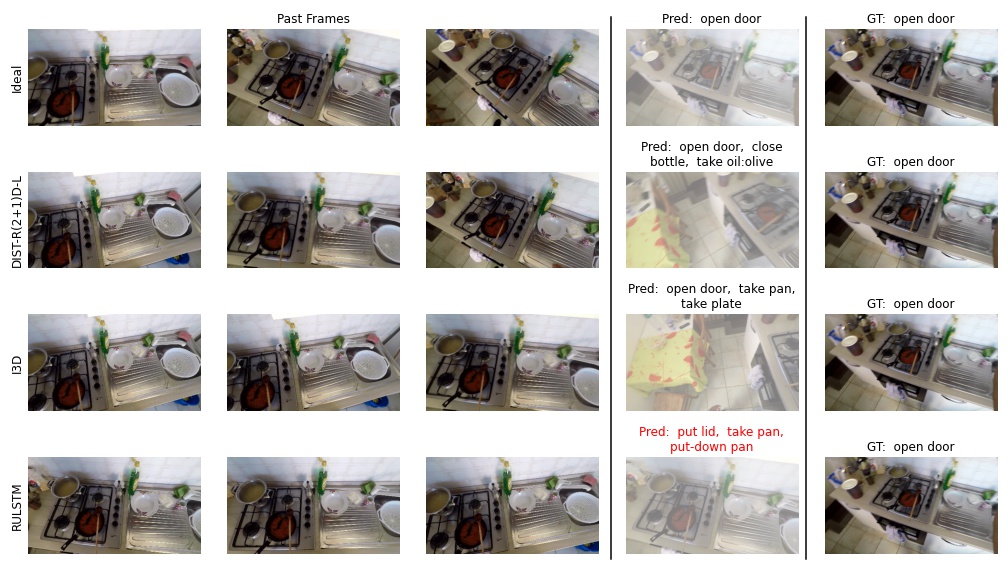}
    \caption{Qualitative examples.}
    \label{fig:qualitative_7}
\end{figure*}

\begin{figure*}
    \centering
    \includegraphics[width=\linewidth]{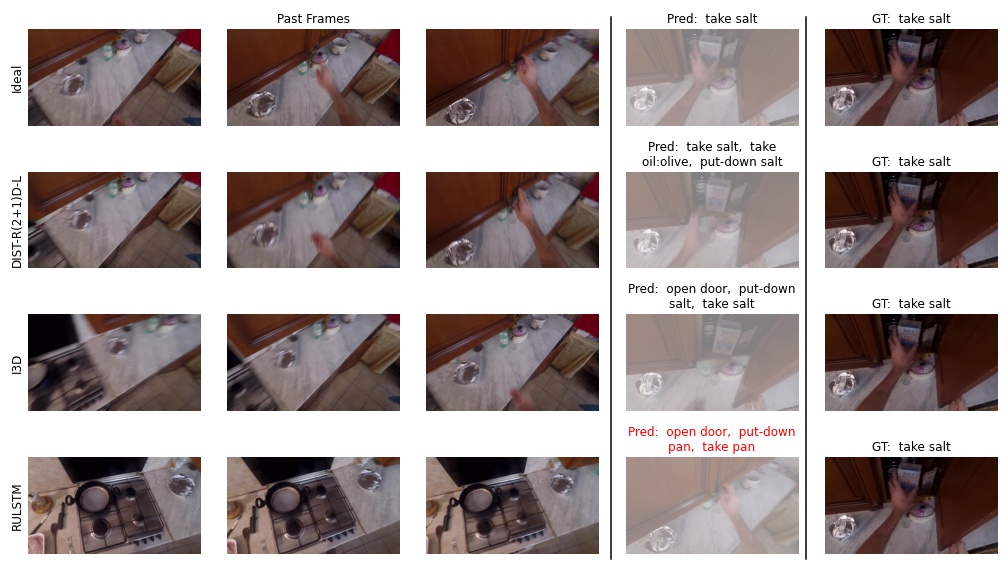}
    \caption{Qualitative examples.}
    \label{fig:qualitative_8}
\end{figure*}

\begin{figure*}
    \centering
    \includegraphics[width=\linewidth]{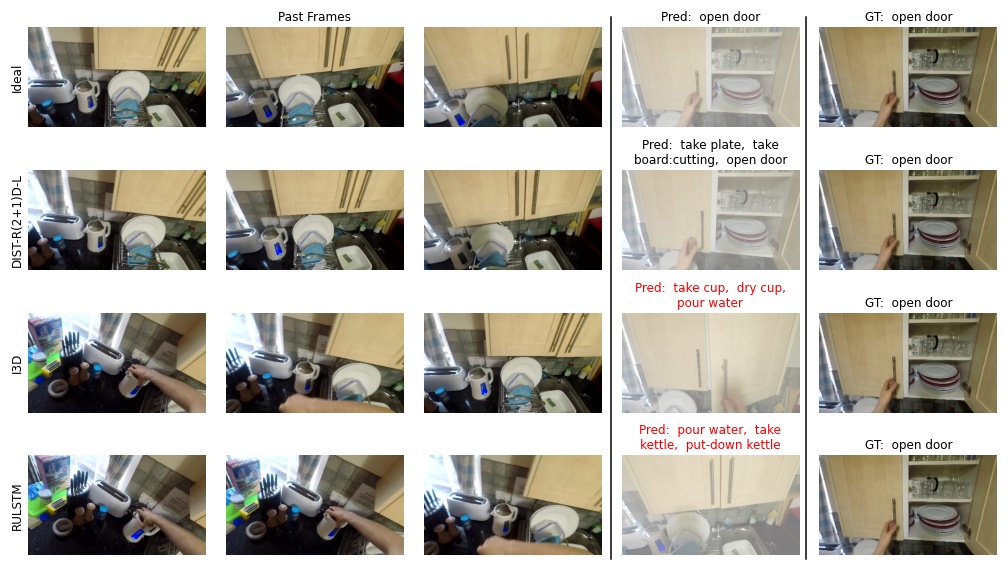}
    \caption{Qualitative examples.}
    \label{fig:qualitative_9}
\end{figure*}

\begin{figure*}
    \centering
    \includegraphics[width=\linewidth]{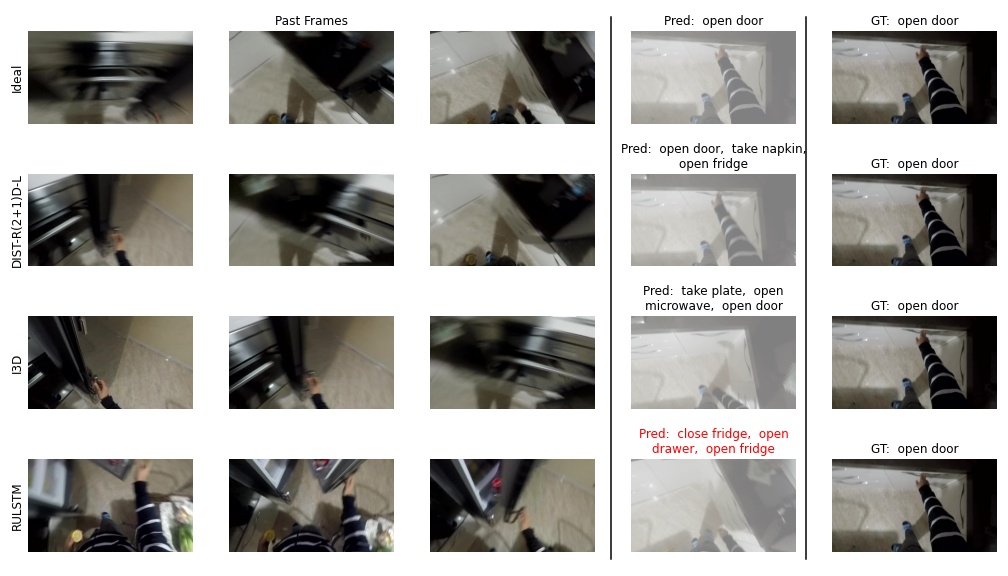}
    \caption{Qualitative examples.}
    \label{fig:qualitative_10}
\end{figure*}

\begin{figure*}
    \centering
    \includegraphics[width=\linewidth]{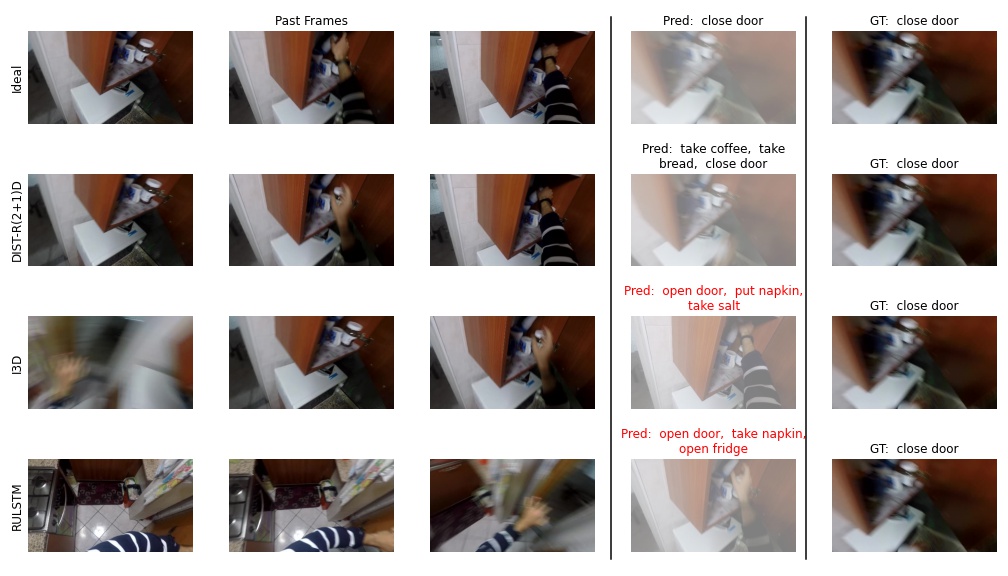}
    \caption{Qualitative examples.}
    \label{fig:qualitative_11}
\end{figure*}

Figures~\ref{fig:qualitative_1}-\ref{fig:qualitative_11} report additional qualitative examples. 
As can be noted, RULSTM usually observes a significantly different input sequence, which makes predictions less effective 

(e.g., Figures~\ref{fig:qualitative_1},~\ref{fig:qualitative_3},~\ref{fig:qualitative_8},~\ref{fig:qualitative_10},~\ref{fig:qualitative_11}).
Despite DIST-R(2+1)D and I3D tend to observe similar inputs, the former tends to make better predictions than the latter (e.g., Figures~\ref{fig:qualitative_2},~\ref{fig:qualitative_9},~\ref{fig:qualitative_11}).

%

{\small
\bibliographystyle{ieee_fullname}
\bibliography{egbib}
}